\pdfoutput=1
\documentclass[letterpaper, 10 pt, conference]{ieeeconf}  %

\IEEEoverridecommandlockouts                              %

\usepackage{amsmath} %
\usepackage{amssymb}  %

\makeatletter
\let\NAT@parse\undefined
\makeatother
\usepackage{graphicx}
\usepackage{booktabs}
\usepackage{textcomp}
\usepackage{caption}
\usepackage{array}
\usepackage[space, compress]{cite}
\usepackage{tabularx}
\usepackage{multirow}
\usepackage{adjustbox}
\let\labelindent\relax
\usepackage{enumitem}
\usepackage{wrapfig}
\usepackage[table]{xcolor}

\definecolor{ourshl}{RGB}{235,245,255}
\definecolor{abup}{HTML}{1B7E3A}
\definecolor{abdown}{HTML}{B33A3A}

\usepackage[normalem]{ulem}

\PassOptionsToPackage{hyphens}{url} %
\usepackage[bookmarks=true]{hyperref}
\title{\LARGE \bf
Learning Force-Regulated Robotic Manipulation \\with a Low-Cost Tactile-Force-Controlled Gripper
}

\author{%
Xuhui Kang$^{*1}$, Tongxuan Tian$^{*1}$, Sung-Wook Lee$^{1}$, Binghao Huang$^{2}$, Yunzhu Li$^{2}$, Yen-Ling Kuo$^{1}$\\[0.6ex]
$^{1}$University of Virginia \qquad $^{2}$Columbia University\\[0.3ex]
{\tt\small \{xuhui, nua3jz, dcs3zc, ylkuo\}@virginia.edu \quad \{binghao.huang, yunzhu.li\}@columbia.edu}\\[0.3ex]
{\small $^{*}$Equal contribution}%
}

\begin{document}

\bstctlcite{IEEEexample:BSTcontrol}

\maketitle
\thispagestyle{plain}
\pagestyle{plain}

\begin{abstract}
Successfully manipulating many everyday objects, such as potato chips, requires precise force regulation.
Failure to modulate force can lead to task failure or irreversible damage to the objects.
Humans can precisely achieve this by adapting force from tactile feedback, even within a short period of physical contact.
We aim to give robots this capability.
However, commercial grippers exhibit high cost or high minimum force, making them unsuitable for studying force-controlled policy learning with everyday force-sensitive objects.
We introduce TF-Gripper, a low-cost (\textasciitilde\$150) force-controlled parallel-jaw gripper that integrates tactile sensing as feedback.
It has an effective force range of 0.45--45~N and is compatible with different robot arms.
Additionally, we designed a teleoperation device paired with TF-Gripper to record human-applied grasping forces.
{The gripper hardware and teleoperation system serve as enabling tools for controlled force-regulated manipulation studies.}
While we can train standard low-frequency policies with the collected force data, achieving reliable performance remains challenging due to the reactive and contact-dependent nature of force-regulated manipulation.
To overcome this, we propose RETAF (REactive Tactile Adaptation of Force), a framework that decouples grasping force control from arm pose prediction.
RETAF regulates force at high frequency using wrist images and tactile feedback, while a base policy predicts end-effector pose and gripper open/close action.
Our experiments show that, compared to position control, direct force control with TF-Gripper improves grasp stability and overall task performance across six real-world tasks.
We further show that tactile feedback is essential for force regulation, and that RETAF consistently outperforms baselines and can be integrated with various base policies.
We hope this work opens a path for scaling the learning of force-controlled policies in robotic manipulation.
Project page: \url{https://force-gripper.github.io/}.

\end{abstract}

\section{Introduction}
\begin{figure}[tbp]
    \centering
    \includegraphics[width=1.0\linewidth]{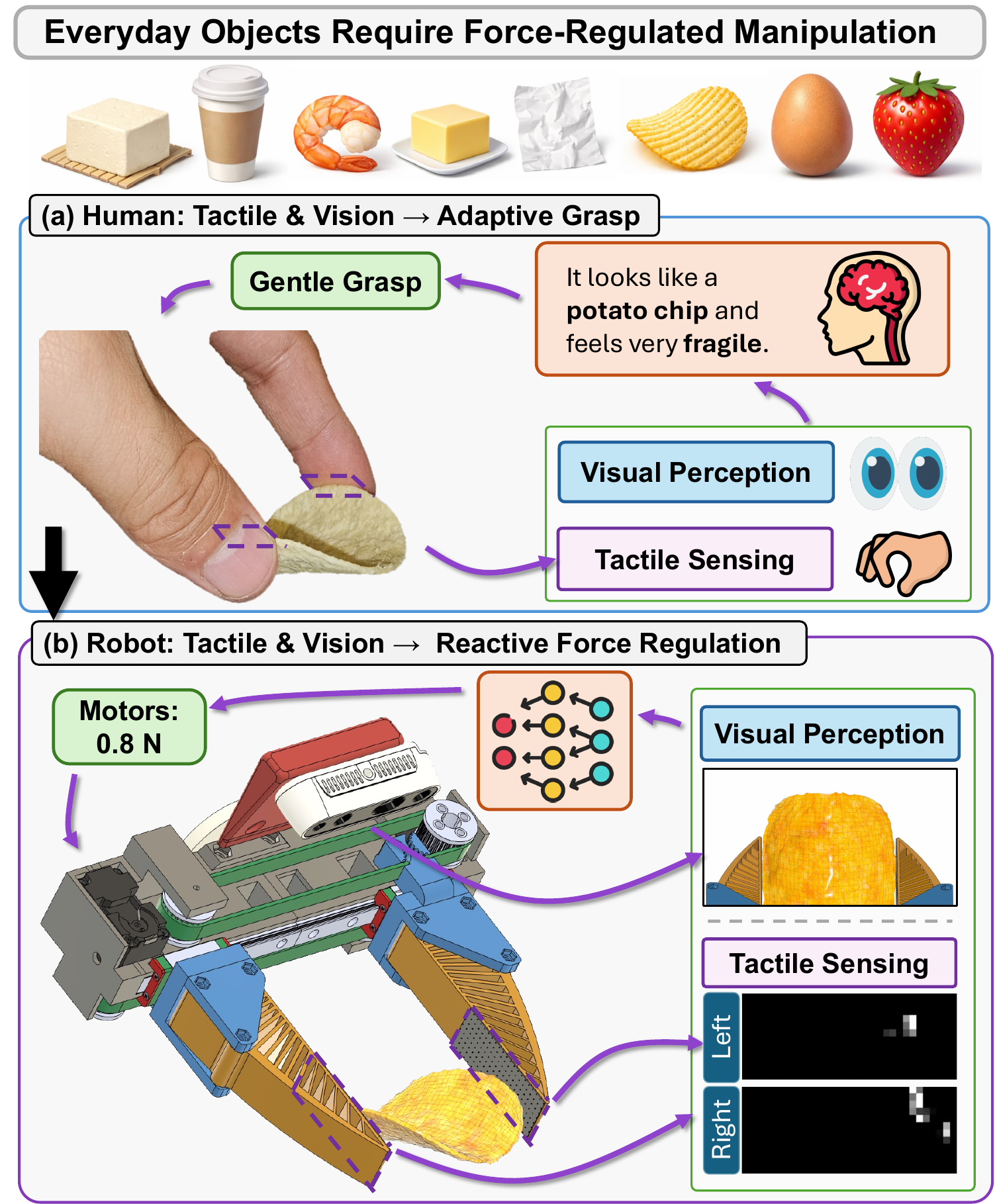}
    \caption{To mimic the human ability to regulate force through tactile feedback to manipulate everyday objects, we develop a force-controlled gripper with tactile sensing for learning reactive and gentle manipulation policies.
    }
    \label{fig:human_grasp}
\end{figure}

\begin{figure}[!htbp]
    \centering
    \includegraphics[width=1.0\linewidth]{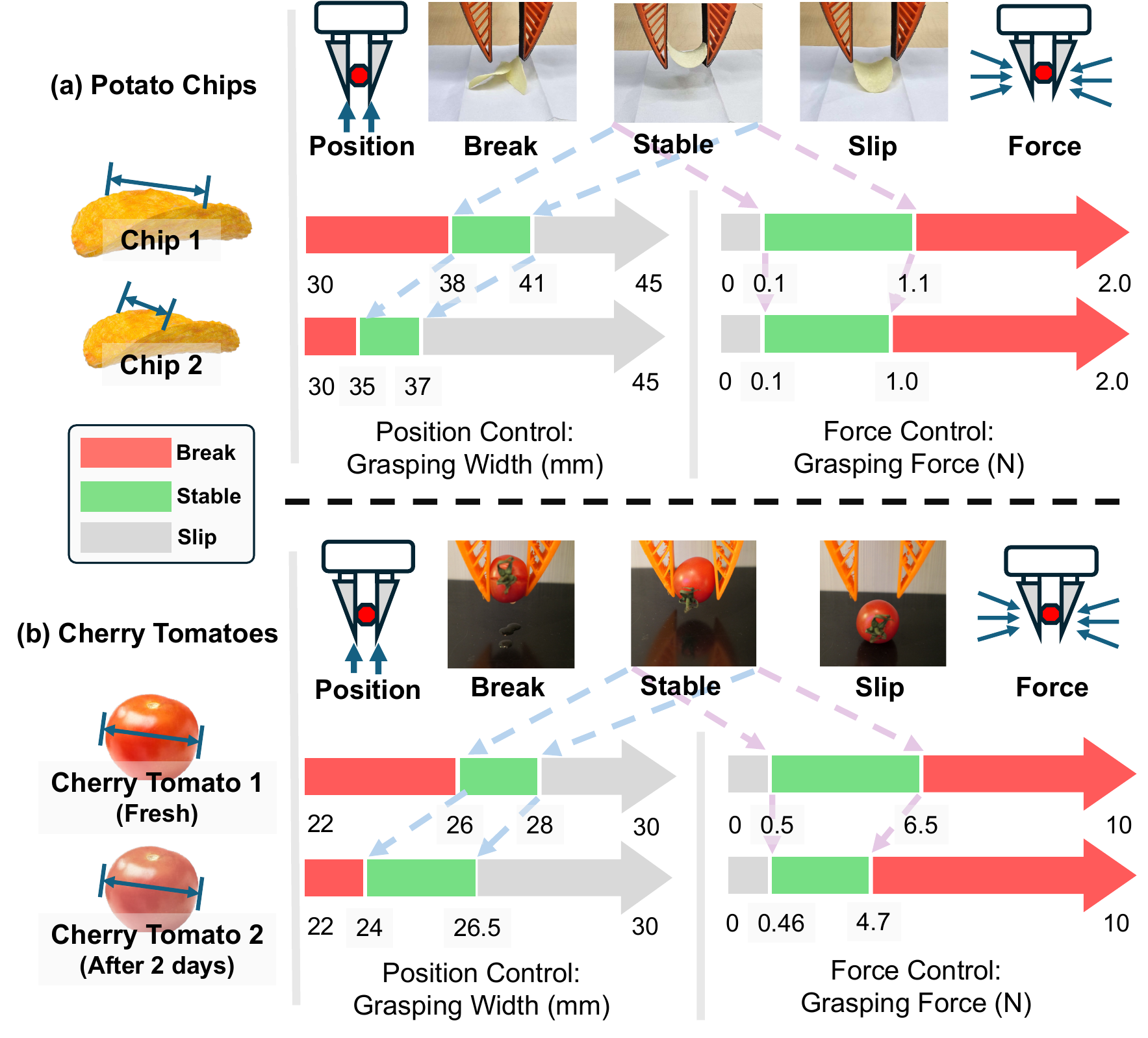}
    \caption{Gripper position vs. force control on objects with similar appearance but different physical properties.
    (a) The same type of potato chips but with a small variation in size.
    (b) The same cherry tomato after two days of softening.
    }
    \label{fig:introduction}
\end{figure}

Humans have demonstrated strong capability in manipulating objects that require precise force regulation, such as picking up potato chips or squeezing ketchup from bottles.
We achieve this by adjusting forces from tactile feedback even within a brief moment of contact~\cite{johansson2009coding}, see Fig.~\ref{fig:human_grasp}(a).
With insufficient force, the task could not proceed, but excessive force could lead to irreversible object damage.
However, existing works on robotic systems, with or without tactile feedback~\cite{yuan2017gelsight,gelsight2022mini,huang20243dvitac}, mostly simplify gripper control to binary open/close actions or gripper position/width prediction~\cite{oxe2024open,zhu2025touch,xu2025exumi,liu2025vitamin,mees2022calvin,wu2025freetacman,tian2025o3afford}.
Such position control can be viewed as an indirect proxy for force control, as the manipulation goal is ultimately to apply an appropriate contact force.
However, using position control to generate appropriate force is highly object- and task-dependent.
For example, in Fig.~\ref{fig:introduction}(a), a small variation in sizes can yield significant differences in stable grasping width range, whereas the stable grasping force range remains similar and can lead to more robust policies.
In Fig.~\ref{fig:introduction}(b), on the other hand, the appearance of cherry tomatoes is almost identical, but it requires a different width range to generate sufficient friction for a stable grasp.
To successfully pick up the cherry tomato, the force control can adjust the force range from the tactile feedback, whereas position control will need to integrate information about the perceived tomato size and tactile input to infer the gripper width, which is harder to learn.

To study how the robot can effectively learn the human-like tactile-force capability, we need a gripper that supports force control.
Many commercial grippers provide force control, but they are often expensive, difficult to adapt across different robots, and offer a high minimum force that is not well suited for delicate, soft, and fragile everyday objects, as summarized in Table~\ref{tab:force_control_grippers}.
To enable more research in learning force-regulated manipulation, we develop and open-source TF-Gripper (\textbf{T}actile-\textbf{F}orce-Controlled Gripper), a universal and low-cost robotic gripper that provides fine-grained force regulation with tactile sensing, see Fig.~\ref{fig:human_grasp}(b).
The TF-Gripper includes a teleoperation device that captures the force applied by a human operator through a spring-like actuator, which provides local kinesthetic feedback while estimating the applied force via motor current, and maps this signal to the gripper mounted on the robot during human demonstrations.
This setting is sufficient to collect high-quality force control data, as the compliant interaction reduces variability while preserving fine-grained force modulation for learning force-regulated manipulation.

\begin{table}[h]
\caption{Mainstream Grippers with Force Control}
\label{tab:force_control_grippers}
\begin{adjustbox}{width=\linewidth}
\begin{tabular}{lcllc}
\midrule
\textbf{\begin{tabular}[c]{@{}l@{}}Gripper \end{tabular}} &
\textbf{\begin{tabular}[c]{@{}c@{}}Cross-Robot \\ Compatibility\end{tabular}} &
\textbf{\begin{tabular}[l]{@{}l@{}}Force \\ Range (N)\end{tabular}} &
\textbf{\begin{tabular}[l]{@{}l@{}}Travel \\ Range (mm)\end{tabular}} &
\textbf{Price} \\ \midrule
Franka Hand~\cite{franka2020hand}      & No  & 30--70  & 80  & \textasciitilde\$2,000 \\
Robotiq 2F-85~\cite{robotiq2019gripper}    & Yes & 20--235 & 85  & \textasciitilde\$5,000 \\
xArm Gripper G2~\cite{ufactoryUFACTORY}  & No  & 10--50  & 84  & \textasciitilde\$2,200 \\
WSG-50~\cite{weissroboticsSeriesSeries}           & Yes & 5--80   & 110 & \textasciitilde\$5,600 \\
OnRobot RG2~\cite{onrobot_rg2}      & Yes & 3--40   & 110 & \textasciitilde\$5,500 \\
Flexiv Grav~\cite{flexivFlexivGrav}      & No  & 1--100  & 100 & \textasciitilde\$6,000  \\
\textbf{TF-Gripper} (Ours)             & Yes & 0.45--45   & 95  & \textasciitilde\$150   \\ \midrule
\end{tabular}
\end{adjustbox}
\end{table}

It is possible to train a force control policy using the existing model architecture like diffusion policy (DP)~\cite{chi2023diffusion} or vision-language-action (VLA) models~\cite{kim24openvla,black2024pi_0}.
These models jointly predict the robot arm pose and gripper command (mostly, gripper width in prior works) at a relatively low frequency (e.g., 1--10~Hz).
However, force control introduces a fundamentally different control objective from position control.
Rather than reaching a gripper position, the gripper closes by continuously applying a specified force, and success depends on how that force is regulated during a short, contact-rich interaction phase.
A low-frequency policy is insufficient to capture fine-grained contact dynamics and thus struggles to support force-regulated manipulation which requires fast, contact-driven adaptation.
Moreover, directly fusing tactile inputs with global visual observations can confuse the policy and often leads to unstable learning~\cite{mees2022calvin,xue2025reactive}.

To address these issues, we propose \textbf{RETAF} (\textbf{RE}active \textbf{T}actile \textbf{A}daptation of \textbf{F}orce), a policy design that explicitly decouples arm pose prediction from grasping force prediction.
RETAF employs a force adaptation policy to reactively predict force at a high frequency ($>$30~Hz) based on wrist-view images and tactile feedback, while a base policy predicts end-effector pose and gripper open/close actions.
This design enables rapid force adaptation during contact and only considers tactile information when needed.
We evaluate TF-Gripper and RETAF with six real-world manipulation tasks that require precise force regulation to complete.
We collected 50 human demonstrations for each task using the TF-Gripper.
Experimental results show that RETAF can be integrated with different pose prediction policies and consistently improves grasp stability, pose execution quality, and overall task success in force-regulated manipulation.

In summary, our work makes the following contributions:
\begin{enumerate}[leftmargin=*, labelsep=0.5em]
\item {We introduce TF-Gripper, a low-cost (\textasciitilde\$150) tactile-force-controlled gripper paired with a force-aware teleoperation device for learning force-regulated manipulation.}
\item We propose RETAF, a policy framework that decouples force control from end-effector pose prediction and can be integrated with any existing pose-prediction policies.
\item We evaluate RETAF on force-regulated manipulation tasks, showing improved reactive force regulation over position control.
\end{enumerate}

\section{Related Work}

\noindent \textbf{Force-Controlled Grippers.}
Force-controlled grippers support compliant manipulation, typically through open-loop current/torque estimates, as in Robotiq 2F-85~\cite{robotiq2019gripper}, WSG-50~\cite{weissroboticsSeriesSeries}, and MAGPIE~\cite{correll2024versatile}, or closed-loop force/contact feedback, such as fingertip torque sensing in WSG-50 and ECHO~\cite{bazhenov2025echoopensourcelowcostteleoperation} and tactile force estimation in FARM~\cite{helmut2025tactile}.
Open-loop designs can suffer from backlash and position-dependent moment arms, while closed-loop designs often require extra sensors or specialized fingertips that reduce compatibility with general-purpose~\cite{chi2024universal} or task-customized fingertips~\cite{ha2021fit2form}.
Our gripper instead uses a timing-belt transmission to reduce backlash and maintain a consistent moment arm, enabling low-cost~\cite{shi2025toddlerbot} open-loop force regulation with tactile feedback for learning-based adaptation.

\noindent \textbf{Force-Aware Teleoperation and Data Collection.}
Learning force-regulated manipulation requires demonstrations with grasping force, but existing teleoperation frameworks~\cite{iyer2024open,fu2024mobile} and data collection systems~\cite{chi2024universal} mainly record pose and gripper open/close or width.
Recent methods infer force from joystick and motor feedback~\cite{bazhenov2025echoopensourcelowcostteleoperation} or tactile observations~\cite{helmut2025tactile}; FTF~\cite{adeniji2025feel} measures 3D contact forces with a tactile glove, and UMI-FT~\cite{choi2026inthewildcompliantmanipulationumift} augments a gripper with a force sensor.
In contrast, our teleoperation device directly measures the operator's force, maps it to the robot gripper in real time, and pairs with VR controllers or other pose input devices without proxy inference or post-hoc estimation.

\noindent \textbf{Learning with Force and Tactile.}
Early work regulates grip force from tactile signals with hand-designed controllers, e.g., by detecting slip~\cite{romano2011human,dong2019maintaining}, and learns from vision and touch to predict grasp outcomes and adjust grasps~\cite{calandra2018more}.
Tactile sensing improves contact awareness and robustness in learning-based manipulation, with prior work on tactile encoding and policy integration for contact-rich interactions~\cite{huang2025tactile,luu2025manifeel}, reactive control~\cite{she2021cable,xue2025reactive,xu2024letac}, grasping of deformable objects~\cite{han2025learning}, sim-to-real refinement~\cite{huang2025vtrefine}, and tactile and multimodal representation learning~\cite{lee2019making,xu2025unit,zhu2025touch}.
Recent force-aware methods estimate contact force distributions from tactile observations in FARM~\cite{helmut2025tactile} or learn slip-aware force control with a physics-informed energy-based abstraction~\cite{kuo2025tracing}, but they largely use tactile feedback as an auxiliary observation or focus on force estimation.
In contrast, RETAF decouples force regulation from pose prediction and uses high-frequency tactile feedback for reactive force adaptation during contact.

\section{TF-Gripper Hardware System Design}
\begin{figure}[!t]
    \centering
    \includegraphics[width=1.0\linewidth]{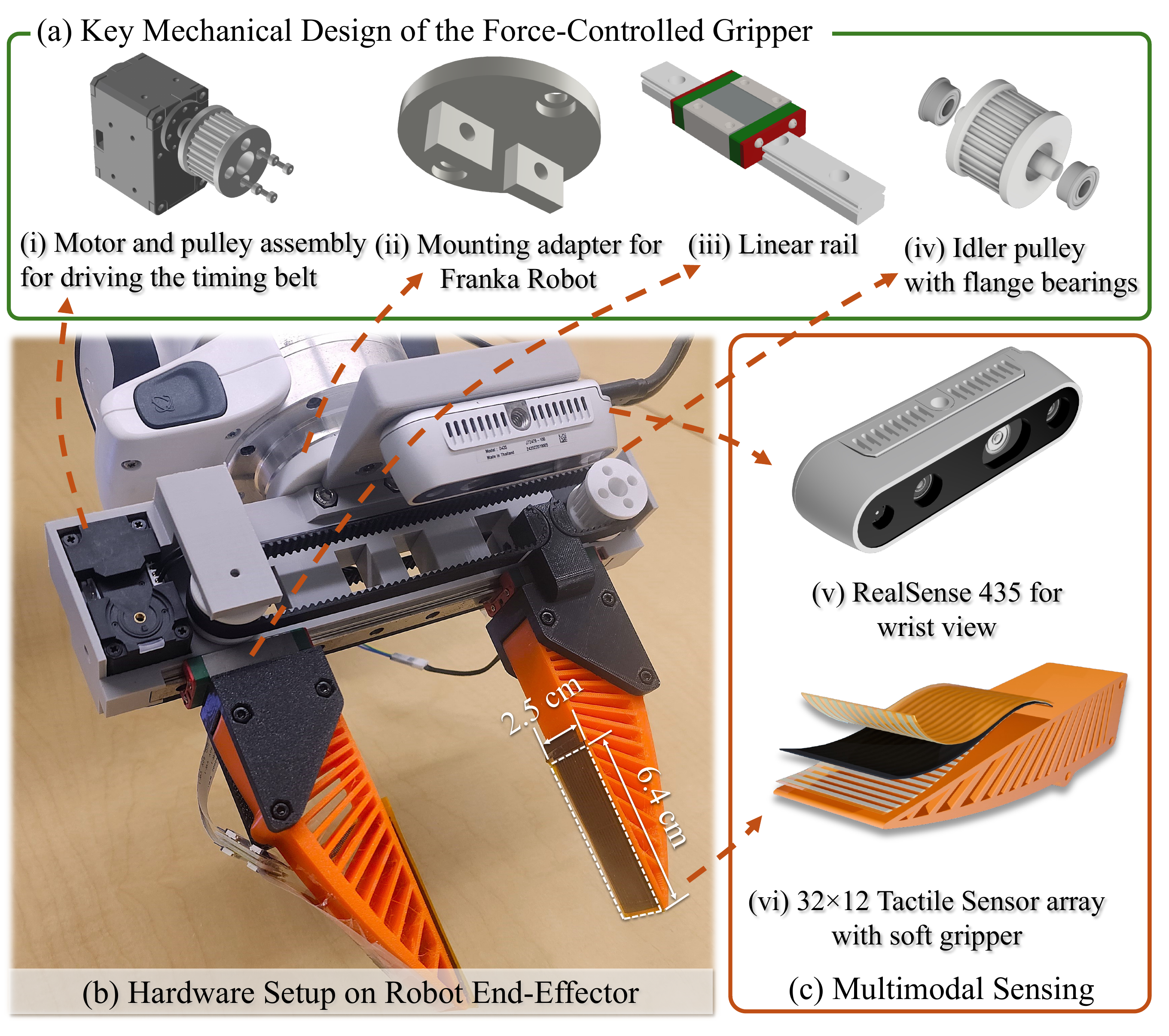}
    \caption{
    Hardware design of the TF-Gripper.
    (a) An adapter connects the gripper to a robot and two motors actuate the fingertips along linear rails by pulling a timing belt.
    (b) The overall TF-Gripper setup.
    (c) A camera provides a wrist view, with soft fingertips integrated with tactile sensors.
    }
    \label{fig:hardware_design}
\end{figure}
\subsection{Gripper Design}
To enable precise, low-cost force control for force-regulated manipulation, we develop \textbf{TF-Gripper}, a parallel-jaw tactile-force-controlled gripper driven by two Dynamixel XL430-W250-T~\cite{robotisxl430} actuators.
As shown in Fig.~\ref{fig:hardware_design}, each motor drives a pulley with timing-belt transmission to actuate one fingertip; the pulley radius defines the effective moment arm, providing a repeatable mapping from motor torque to fingertip force while maintaining mechanical robustness.
The main gripper frame is fully 3D printed.
Each fingertip is mounted on a linear slider that moves along a low-friction linear rail, ensuring stable and smooth parallel motion during grasping (Fig.~\ref{fig:hardware_design}~(iii)).
On the passive side, the timing belt is routed through an idler pulley composed of a cylindrical shaft and two flanged bearings, which reduces friction and improves motion consistency (Fig.~\ref{fig:hardware_design}~(iv)).
To support compliant contact with objects, the fingertips are equipped with soft fingertip pads from UMI~\cite{chi2024universal} (Fig.~\ref{fig:hardware_design}~(vi)).
To further enable reliable tactile perception at low cost, we integrate thin and soft piezo-resistive tactile sensors, \textit{FlexiTac}~\cite{huang20243dvitac,zhu2025touch}, which capture both contact area geometry and force-related information while maintaining mechanical softness.
The gripper is mounted on the robot via an interchangeable adapter.
For the Franka Research 3 and UR5, we provide an ISO 9409-1-A50 adapter (Fig.~\ref{fig:hardware_design}~(ii)).

The Dynamixel actuators exhibit a near-linear current-to-force relationship when evaluated 1~cm from the end of the fingertip.
This enables accurate and stable force control over a range of approximately 0.45--45~N, with a command resolution of about 0.2~N.
\begin{figure}[!htbp]
    \centering
    \includegraphics[width=1.0\linewidth]{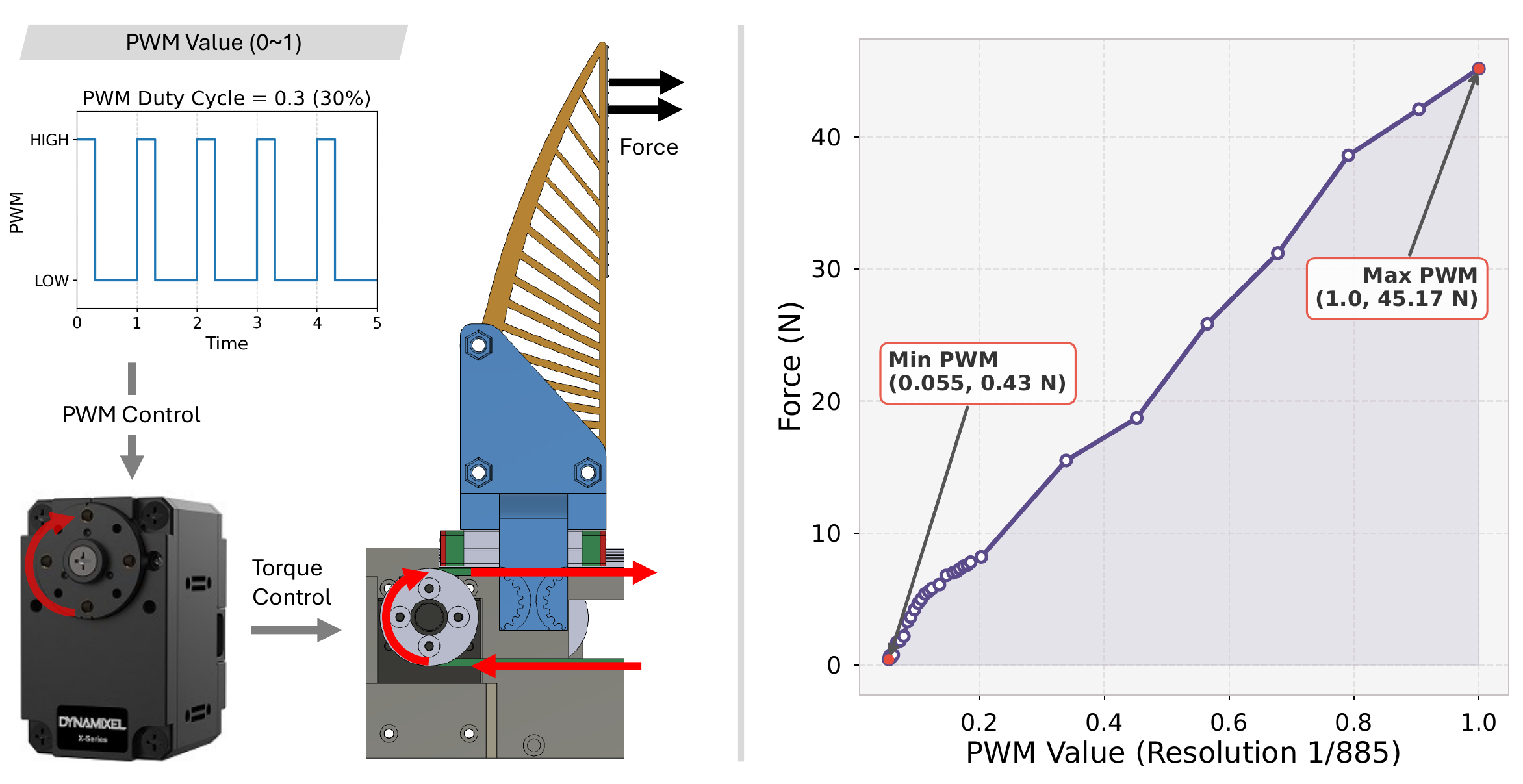}
    \caption{{PWM--force actuation of the TF-Gripper. \textbf{Left:} the PWM command controls the motor rotational torque, which drives the fingertip through the pulley and timing-belt transmission to regulate the applied contact force. \textbf{Right:} the measured PWM--force relationship at the fingertip contact surface, showing the effective grasping force.}}
    \label{fig:pwm_force}
\end{figure}
{Fig.~\ref{fig:pwm_force} further illustrates the full actuation chain from the PWM to motor torque and then to fingertip contact force, together with the measured PWM--force calibration at the fingertip.}
We further evaluate force stability under temperature variations.
After running the actuator at 20\% current for 20 minutes at 25\,$^\circ$C, we observe an average force drift of about 9\%.
Using the motor's built-in temperature sensor for compensation reduces this error to below 4\%, indicating stable force control.

{The PWM--force mapping is calibrated directly with a force gauge at the fingertip, so static friction is already included and no motor-current-based force estimate is involved.
To quantify the remaining nonlinearity, belt hysteresis, we measure the fingertip force with the force gauge at commanded forces of 1, 5, and 10~N during loading and unloading, with 10 repetitions each.
With a loading-only calibration, unloading shows a hysteresis error of up to 1.0~N (19\%) at 5~N.
We therefore calibrate both directions with the force gauge, after which the mean error is within 7\% at all setpoints, on par with the $\pm$10\% force repeatability specified for the Robotiq 2F-85~\cite{robotiq2019gripper} (Appendix~\ref{app:force_accuracy}).
The lower bound of the force range (about 0.45~N) is the deadband visible in the calibration, below which the motor torque cannot overcome the transmission friction (Fig.~\ref{fig:pwm_force}).
}

\subsection{Teleoperation Device}

Most teleoperation methods, such as VR controllers, SpaceMouse devices, or kinesthetic teaching, focus on joint or end-effector pose control and limit gripper commands to open/close or width control, making human-applied grasping force difficult to capture.
To collect demonstrations with explicit force control, we design a low-cost teleoperation interface (\textless\$100) that mirrors the gripper's actuation mechanism.
\begin{figure}[!htbp]
    \centering
    \includegraphics[width=1.0\linewidth]{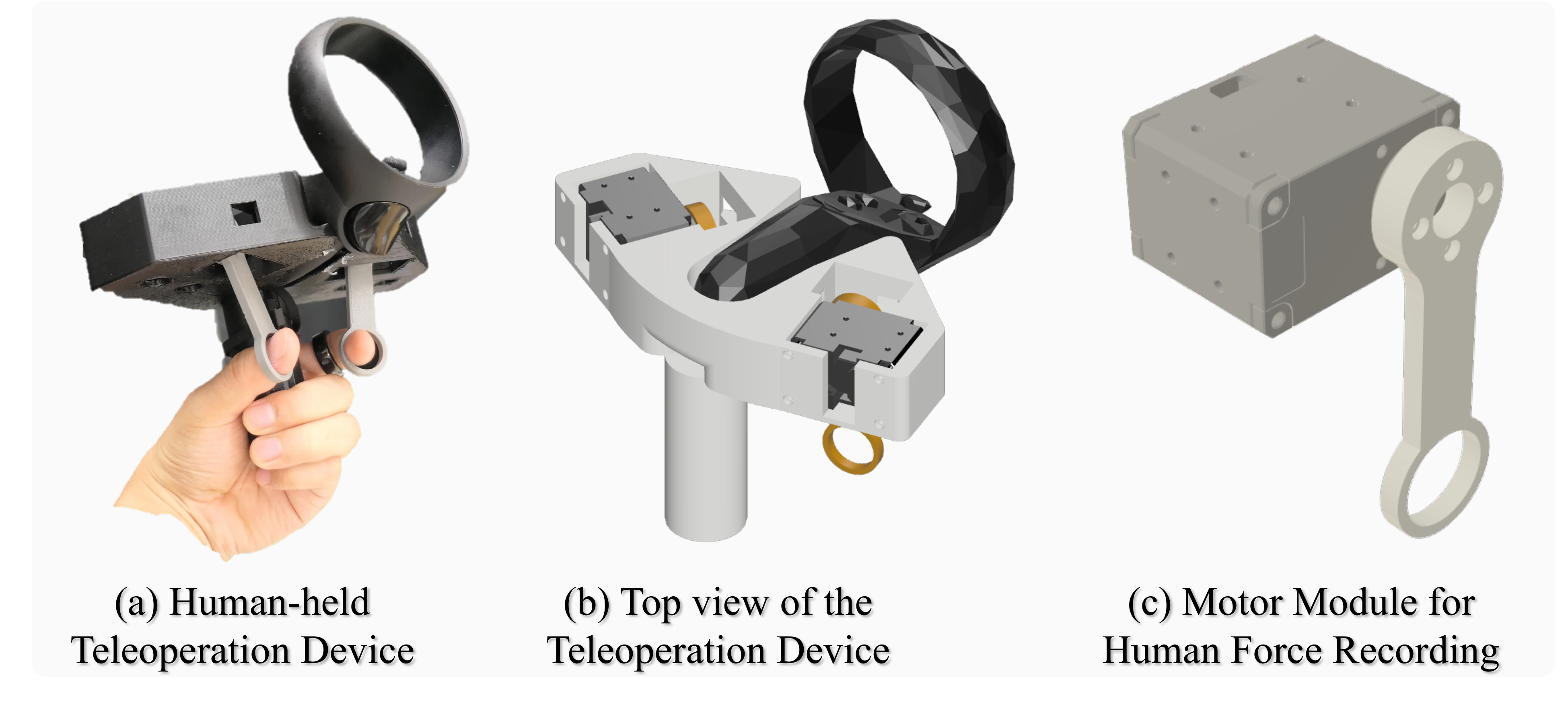}
    \caption{Design of the teleoperation device. A VR controller controls the robot's end-effector pose. Two finger rings are connected to the motors; they detect the force applied by the human operator to drive the TF-Gripper.}
    \label{fig:dat_collection}
\end{figure}
As shown in Fig.~\ref{fig:dat_collection}, the device uses a VR controller for end-effector pose and two finger rings connected to the same Dynamixel actuator as the gripper for force input.
The actuator provides spring-like resistance for local kinesthetic feedback while measuring the operator's force via motor current; this calibrated signal is mapped to the robot gripper current command in real time.

\section{RETAF: REactive Tactile Adaptation of Force}

\subsection{Problem Formulation}
We formulate a robotic manipulation task as a sensorimotor problem.
At time $t$, the observation is $o_t = \{V_t, \mathcal{T}_t\}$, where $V_t$ denotes non-tactile observations (e.g., RGB images, proprioception) and $\mathcal{T}_t$ tactile readings.
A policy $\pi$ maps these observations to an action $a_t = [a^{\text{pose}}_t, a^{\text{grip}}_t] \sim \pi(V_t, \mathcal{T}_t)$, consisting of the arm command $a^{\text{pose}}_t$ (e.g., end-effector pose) and the gripper command $a^{\text{grip}}_t$ (typically position).
This coupled formulation has two mismatches for force-regulated
manipulation.
\textbf{1) Frequency mismatch.}
Stable force regulation must react to transient tactile events (e.g., slip) within a short time ($\tau_{\text{react}} \le 50$\,ms), whereas pose prediction uses large backbones with a latency
of $100$--$1000$\,ms.
A coupled policy is bottlenecked by this slow inference ($\tau_{\text{lat}} \gg \tau_{\text{react}}$
).
\textbf{2) Learning with tactile input.}
Tactile signals are sparse and contact-dependent, whereas visual observations are dense.
Fusing them directly often causes unstable learning, as the policy must weigh the transient tactile signal against the dominant visual context~\cite{mees2022calvin,xue2025reactive}.

\begin{figure}[!htbp]
\centering
\includegraphics[width=1.0\linewidth]{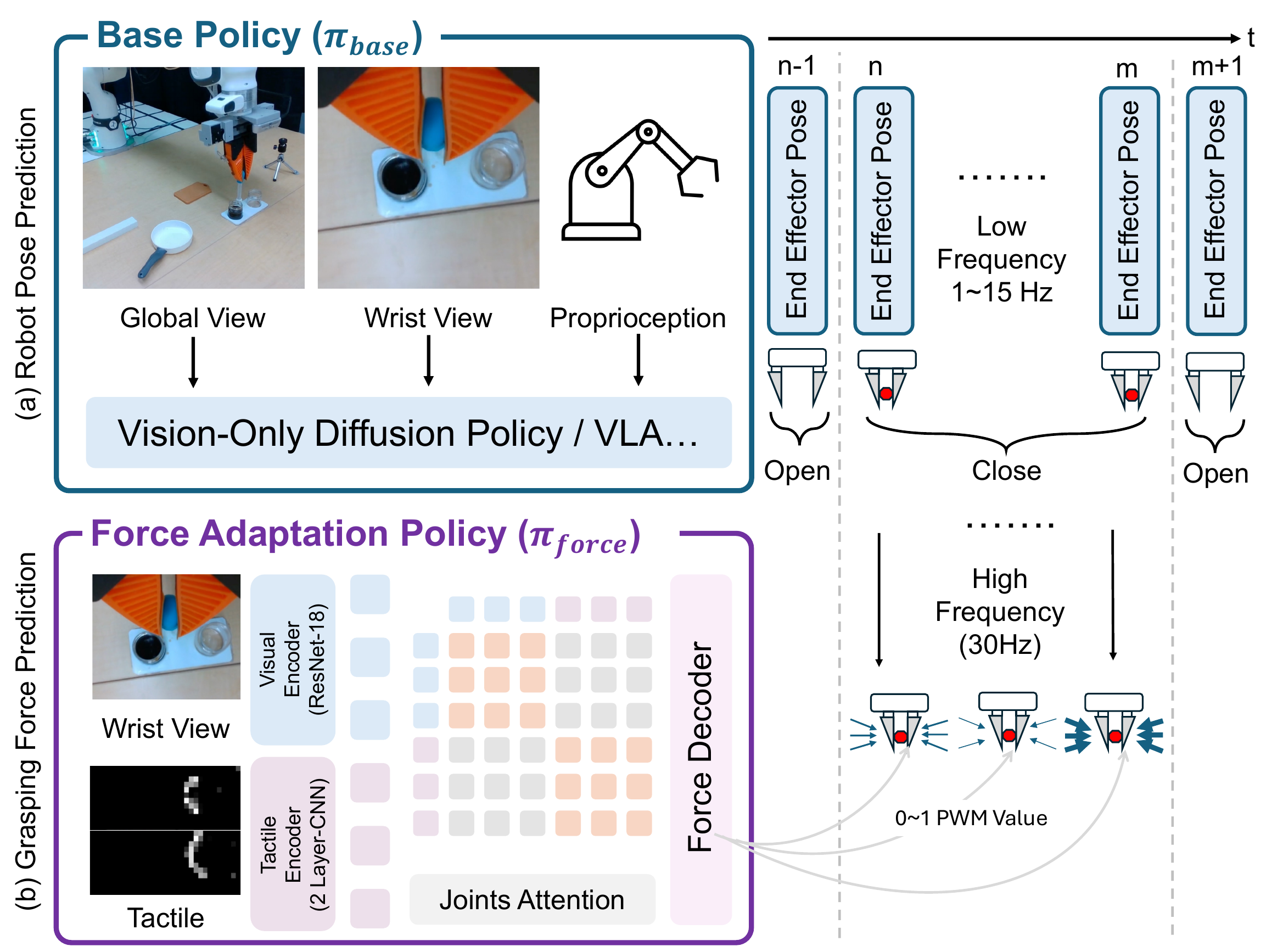}
\caption{
Overview of the RETAF framework.
The base policy predicts end-effector pose and gripper open/close from visual and proprioceptive observations at a low frequency.
When the gripper closes, the force adaptation policy is activated to predict grasping force at a high frequency using joint attention over wrist-view and tactile inputs.
}
\label{fig:our_method}
\end{figure}

\subsection{The RETAF Framework}
To address these challenges, we decouple the policy into a Base Policy $\pi_{\text{base}}$ for trajectory generation and a Force Adaptation Policy $\pi_{\text{force}}$ for contact
regulation (Fig.~\ref{fig:our_method}).
The force policy is activated by a discrete gripper action $k_t \in \{\text{open}, \text{close}\}$ predicted by the base policy (Appendix~\ref{app:gripper_action} studies a variant without this prediction).

\noindent \textbf{Base Policy ($\pi_{\text{base}}$).}
We use existing visuomotor policy architectures without
modification.
Operating at low frequency, $\pi_{\text{base}}$ predicts the arm pose and the discrete gripper open/close from non-tactile observations, $(a^{\text{pose}}_t, k_t) = \pi_{\text{base}}(V_t)$,
and is trained via behavior cloning on expert actions $a^*_t = (a^{\text{pose}*}_t, k^*_t)$
with the architecture-specific loss (e.g., the noise-prediction loss of Diffusion Policy).

\noindent \textbf{Force Adaptation Policy ($\pi_{\text{force}}$).}
The gripper action is gated by $k_t$: $\pi_{\text{force}}$ is activated \textit{only} when the base policy signals a grasp ($k_t=\text{close}$):
\begin{equation}
a^{\text{grip}}_t =
\begin{cases}
\pi_{\text{force}}(I^{\text{wrist}}_t, \mathcal{T}_t) & \text{if } k_t = \text{close} \\
\emptyset \text{ (Open)} & \text{if } k_t = \text{open}
\end{cases}
\end{equation}
When activated, $\pi_{\text{force}}$ predicts the continuous target force $f_t$ at high frequency ($>$30\,Hz).
Since it only needs to reason about grasping force, it attends exclusively to the wrist view $I^{\text{wrist}}_t$ and tactile sensing $\mathcal{T}_t$ through a joint-attention layer~\cite{vaswani2017attention}.
The wrist view provides compact, object-centric visual information (object category, shape, and surface appearance), while tactile sensing captures local contact signals such as deformation and incipient slip.
Attending to these two modalities lets RETAF regulate force precisely without being distracted by the global scene.
We train $\pi_{\text{force}}$ by regressing the demonstrator's force $f^*_t$ with the MSE loss $\mathcal{L}_{\text{force}} = \mathbb{E}\big[\| \pi_{\text{force}}(I^{\text{wrist}}_t, \mathcal{T}_t) - f^*_t \|^2\big]$.
This lightweight design runs at 80\,Hz+ on a single RTX 5070 Ti.

\section{Experimental Setup}

We conduct real-world experiments on tasks that require precise force-regulation to evaluate TF-Gripper and RETAF.

\subsection{Hardware and Data Collection}

\begin{figure}[!htbp]
\centering
\includegraphics[width=1.0\linewidth]{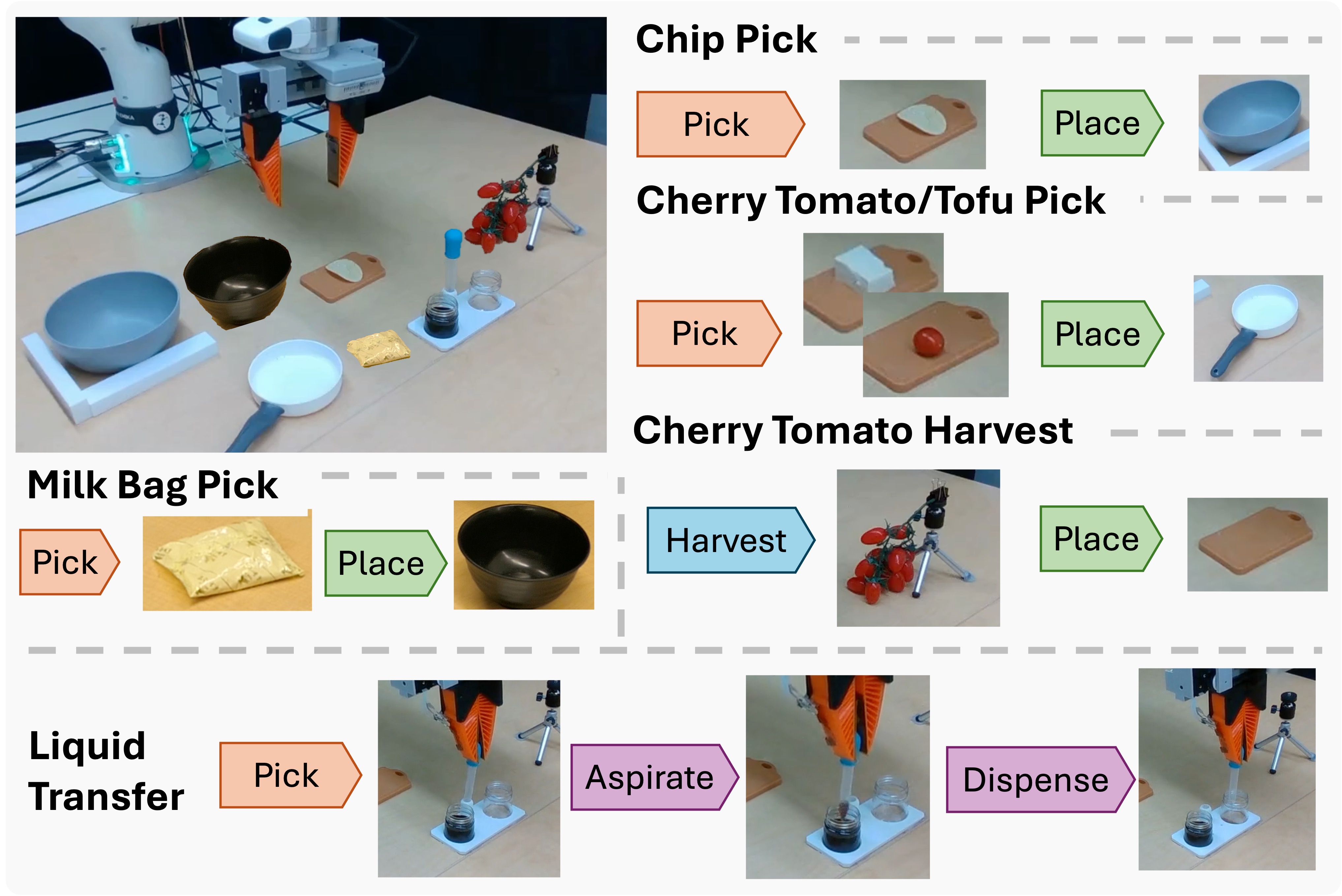}
\caption{
Overview of the evaluation environment and tasks.
}
\label{fig:env_overview}
\end{figure}

\textbf{Hardware Setup.}
Our platform is a Franka Research~3 robot equipped with our TF-Gripper and two RealSense D435i cameras: one third-person view for the global scene and one on the gripper
for the wrist view.
All experiments are conducted on a tabletop with everyday objects such as a plate, a bowl, a pan, a cup, and a
dropper (Fig.~\ref{fig:env_overview}).

\textbf{Data Collection.}
Data is collected with our teleoperation device (Fig.~\ref{fig:dat_collection}) using a Meta Quest~1 as the VR controller.
We record the full trajectory at 15~Hz, including end-effector pose, gripper force, gripper position, and visual and tactile observations.
For each task, we collect 50 demonstrations of 4 to 10 seconds each (examples in Appendix~\ref{app:data_diversity}).
This is on par with prior work in imitation learning and visuo-tactile manipulation~\cite{chi2023diffusion,huang20243dvitac}.
Since the demonstrations explicitly encode force changes, we can compare position control and force control under identical task conditions.

\subsection{Baselines}

We compare RETAF against two learning-based baselines.

\textbf{Diffusion Policy}~\cite{chi2023diffusion} (DP) is one of the most widely adopted manipulation policies and predicts action sequences by
iterative denoising.
We evaluate DP variants with and without tactile input, and with gripper actions represented as either gripper position or grasping force.
Since all policies run on the same TF-Gripper with the same demonstrations and observation space, comparing RETAF with DP w/ Force isolates the algorithmic contribution from the hardware.

\textbf{ViTac-MAE}~\cite{zhu2025touch} extends DP with a tactile encoder pretrained by masked autoencoding.
Comparing it with DP under identical settings tests whether a stronger tactile representation alone is sufficient for precise force control.

\subsection{Implementation Details}

Across all methods, the action space consists of the absolute end-effector position and rotation represented using a 6D rotation representation~\cite{zhou2019continuity}, and a gripper action.
Depending on the method, the gripper action can be either open/close, continuous gripper position, or continuous grasping force.
For visual observations, RGB images are encoded using the image encoder from CLIP~\cite{radford2021learning}.
{The tactile input is the raw tactile image represented as a $12\times32$ matrix.}
We encode the tactile observations using a two-layer convolutional neural network.
All DP and ViTac-MAE variants use a 1D U-Net as the diffusion backbone.
For RETAF, the base policy adopts the same DP architecture to predict only end-effector pose and a binary gripper open/close action using visual observations.
When the gripper closes, the force adaptation policy is activated and predicts the target grasping force from wrist-view images and tactile feedback.
In position-based control experiments, the same force adaptation policy architecture is employed to predict gripper position.

\begin{figure*}[!htbp]
    \centering
    \includegraphics[width=1.0\linewidth]{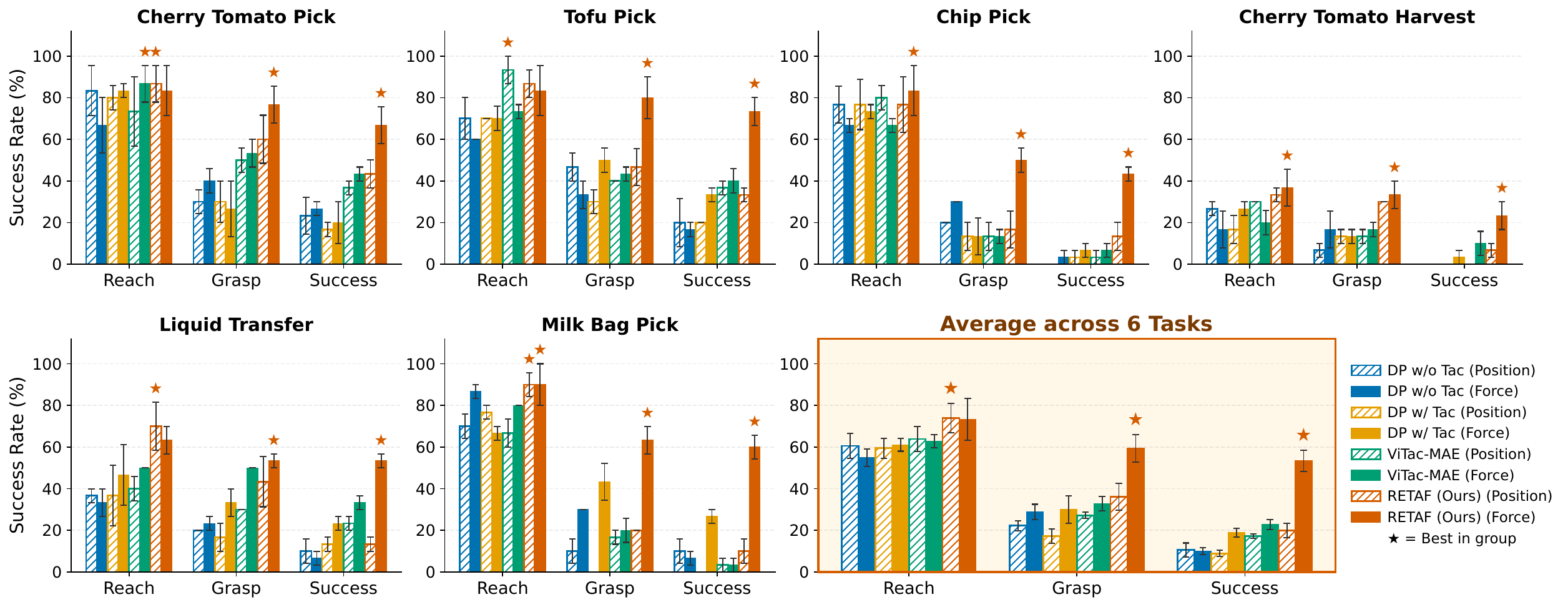}
\caption{
Performance of six manipulation tasks measured in three stages: Reach, Stable Grasp, and Task Success. Bars show the mean $\pm$ standard error over three seeds. Hatched bars use position control and solid bars use force control.
}
    \label{fig:five_mani_tasks_results}
\end{figure*}

\subsection{Tasks Setup}

{We designed six manipulation tasks (see Fig.~\ref{fig:env_overview}) that require continuous and precise grasping force control.
In all tasks, excessive force or overly small gripper width risks breaking objects, while insufficient force or overly wide gripper width leads to slippage.
Table~\ref{tab:task_setup} summarizes the force range and stable-grasp criterion for each task.
For Tofu Pick and Cherry Tomato Pick, deformation is computed from the tactile-contact gripper width and the final stable-grasp width as $(w_{\mathrm{contact}}-w_{\mathrm{stable}})/w_{\mathrm{contact}}$.
Based on measurements of the samples in our experiments, we set the deformation thresholds to $18\%$ for tofu and $23\%$ for cherry tomatoes.
Damage and spilling are judged by visual inspection: any visible crack, fracture, irreversible deformation, liquid spill, or dropped object is counted as a failure.
These criteria were applied consistently across all rollouts and methods. For example, a chip counts as broken only when visible fragments detach from it, while powder shedding at the edges does not.}

{\centering
\captionof{table}{{Task setup and stable-grasp criteria.}}
\label{tab:task_setup}
\scriptsize
\renewcommand{\arraystretch}{0.92}
\setlength{\tabcolsep}{4pt}
\resizebox{\linewidth}{!}{%
\begin{tabular}{@{}lcl@{}}
\toprule
\textbf{Task} & \textbf{Force (N)} & \textbf{Stable-Grasp Criterion} \\
\midrule
Tofu Pick & 2.0--5.0 & No slip/breakage; deformation $<18\%$. \\
Chip Pick & 0.9--1.8 & No slip/crack. \\
Milk Bag Pick & 1.0--3.0 & No slip/leakage. \\
Cherry Tomato Pick & 0.5--7.0 & No slip/damage; deformation $<23\%$. \\
Liquid Transfer & 0.8--3.0 & No slip/drop/spill. \\
Cherry Tomato Harvest & 2.0--5.0 & No slip/damage; deformation $<23\%$. \\
\bottomrule
\end{tabular}%
}
\renewcommand{\arraystretch}{1.0}
}

\subsection{Evaluation Metrics}
{We evaluate each policy with 10 rollouts per task, and repeat the rollouts three times with different random seeds.
}
To better understand why a policy fails, we do not rely on the overall task success rate.
Instead, we decompose each task into three stages to report the average success rate.

\textbf{Reach} measures whether the robot can successfully move to the target object and make contact with it.
This stage is a prerequisite for manipulating objects.

\textbf{Stable Grasp} evaluates whether the object remains stable after grasping, without slipping, dropping, excessive deformation, or damage.

\textbf{Task Success} measures whether the robot can complete the full task such as placing, transferring, or releasing the object as required.

\section{Experiment Results}
We investigate five key research questions:

\begin{enumerate}[leftmargin=*, labelsep=0.5em]
\item \textbf{RQ1.} Does direct force control improve force-sensitive tasks compared to position control?

\item \textbf{RQ2.} Is tactile sensing necessary for precise and effective force control?

\item \textbf{RQ3.} Does RETAF improve force-sensitive tasks compared to non-reactive baselines?

\item {\textbf{RQ4.} How do RETAF's modules affect performance?}

\item {\textbf{RQ5.} How does RETAF generalize to unseen objects and a different robotic platform?}
\end{enumerate}

\begin{figure*}[t]
\centering
\begin{minipage}[c]{0.69\textwidth}
  \includegraphics[width=\linewidth]{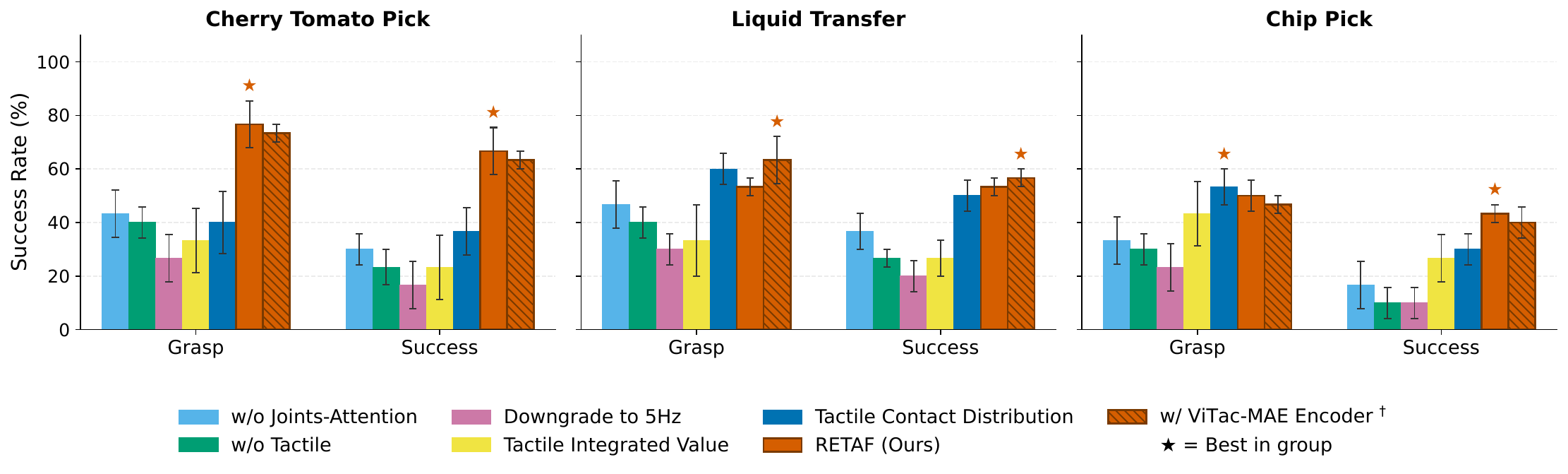}
\end{minipage}\hfill
\begin{minipage}[c]{0.29\textwidth}
{
\centering
\small
\setlength{\tabcolsep}{6pt}
\renewcommand{\arraystretch}{1.25}
\begin{adjustbox}{width=\linewidth}
\begin{tabular}{@{}l cc@{}}
\toprule
Method & Grasp (\%) & Success (\%) \\
\midrule
w/o Joints-Attention & 41.1\,\textcolor{abdown}{$\downarrow$18.9} & 27.8\,\textcolor{abdown}{$\downarrow$26.6} \\
w/o Tactile & 36.7\,\textcolor{abdown}{$\downarrow$23.3} & 20.0\,\textcolor{abdown}{$\downarrow$34.4} \\
Downgrade to 5\,Hz & 26.7\,\textcolor{abdown}{$\downarrow$33.3} & 15.6\,\textcolor{abdown}{$\downarrow$38.8} \\
Tactile Integrated Value & 36.7\,\textcolor{abdown}{$\downarrow$23.3} & 25.6\,\textcolor{abdown}{$\downarrow$28.8} \\
Tactile Contact Distribution & 51.1\,\textcolor{abdown}{$\downarrow$8.9} & 38.9\,\textcolor{abdown}{$\downarrow$15.5} \\
\midrule
\rowcolor{ourshl!40} RETAF (Ours) & \textbf{60.0} & \textbf{54.4} \\
\midrule
w/ ViTac-MAE Encoder$^{\dagger}$ & 61.1\,\textcolor{abup}{$\uparrow$1.1} & 53.3\,\textcolor{abdown}{$\downarrow$1.1} \\
\bottomrule
\end{tabular}
\end{adjustbox}
}

\end{minipage}
\caption{{Ablation study. \textbf{Left:} per-task Grasp/Success rates (mean $\pm$ standard error over three seeds, 10 rollouts each). \textbf{Right:} averages over the 3 tasks (arrows: absolute change vs.\ RETAF). $^{\dagger}$ViTac-MAE: pretrained tactile encoder added to RETAF.}}
\label{fig:ablation}
\end{figure*}

\subsection{Force Control vs.\ Position Control (RQ1)}

We compare force-based and position-based gripper control across all six tasks.
Fig.~\ref{fig:five_mani_tasks_results} shows that force control consistently achieves higher stable grasp rates than position control across different policies, while keeping the performance of the reach stage similar or only slightly lower.
For DP and ViTac-MAE baselines, switching from position to force control also leads to improvements in grasp stability.
For example, DP w/ Tac improves its average stable grasp rate from 17.2\% to 30.0\%, and ViTac-MAE from 27.2\% to 32.8\%.
This trend becomes more pronounced when combined with RETAF.
RETAF with force control achieves an average stable grasp rate of 59.4\%, compared to 36.1\% under position control.
These results indicate that the advantage of force control arises during physical contact rather than object reaching.

\begin{figure}[!htbp]
    \centering
    \includegraphics[width=1\linewidth]{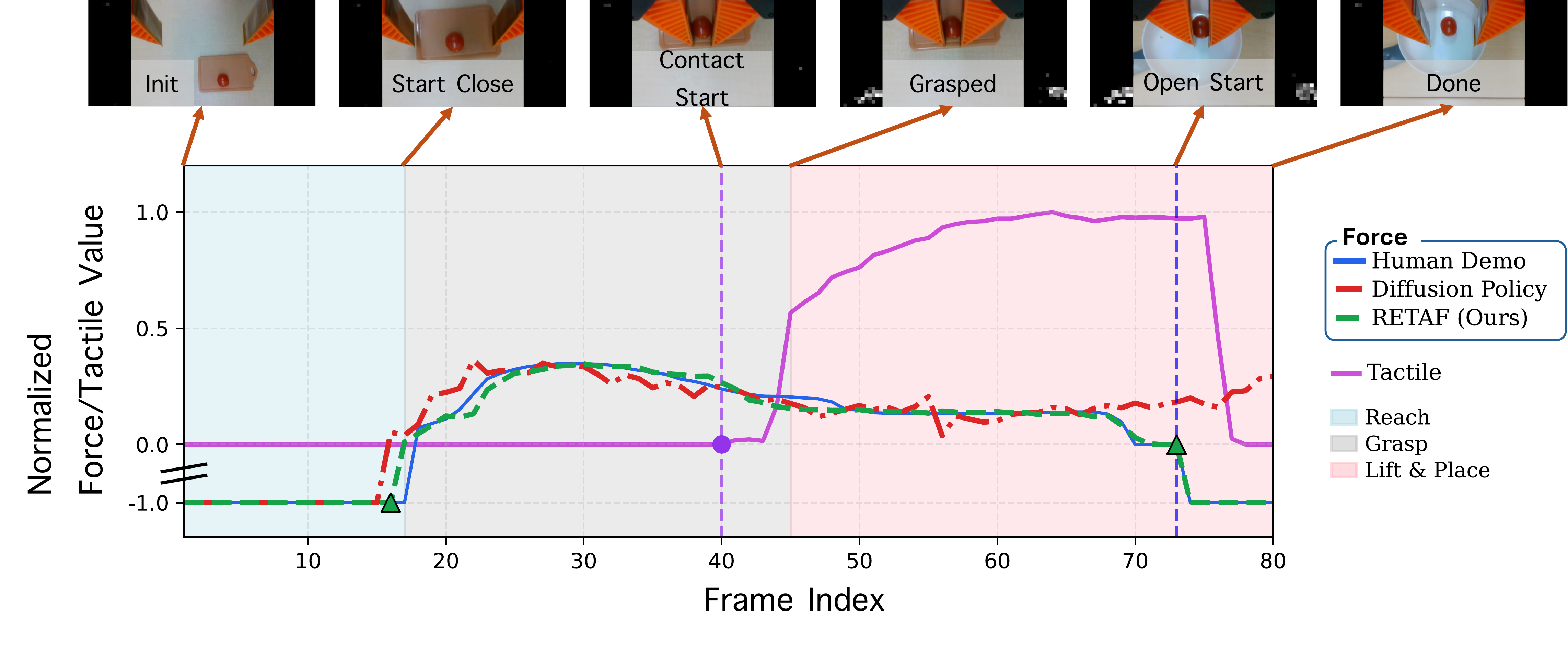}
\caption{
Force prediction on a validation demonstration. For each frame, we compare DP/Force and RETAF predictions with the human-applied force (scaled to $[-1, 1]$, where $-1$ denotes gripper open) and plot the summed tactile value (scaled to $[0, 1]$). RETAF tracks the ground truth and the open timing more closely. Appendix~\ref{app:contact_dynamics} shows more trajectories.
}
    \label{fig:validation}
\end{figure}
\subsection{The Role of Tactile Sensing (RQ2)}
As shown in Fig.~\ref{fig:five_mani_tasks_results}, adding tactile input to DP leads to only small improvements in stable grasp rates compared to the DP without tactile sensing, indicating that directly fusing tactile observations is not sufficient for effective force regulation.
ViTac-MAE achieves better performance by learning stronger tactile representations, resulting in higher grasp stability and task success rate, but its overall improvement remains limited.
In contrast, RETAF explicitly leverages tactile feedback for reactive force adjustment and achieves consistent improvements on average across all stages over all baselines.

Fig.~\ref{fig:validation} provides an example of a validation human demonstration, comparing force predictions from DP and RETAF against the ground-truth human-applied force.
After reaching the target pose, the human applies a larger force to close the gripper, and then adjusts the force upon contact based on tactile and visual feedback.
Notably, the visual appearance before and after contact changes only subtly, making it difficult to infer the required grasping force from vision alone.
Together, these results show that tactile sensing is essential for force-regulated manipulation, and that the way tactile information is used plays a critical role in performance.

\subsection{Ablation of RETAF Modules (RQ3 and RQ4)}
{Fig.~\ref{fig:ablation} shows how RETAF's module designs affect performance. Using only an integrated tactile value lowers the average grasp rate by 23.3\%. Using contact-distribution features also reduces performance because these compact representations discard object-specific contact information, e.g., contact shape and spatial tactile values. Using a pretrained ViTac-MAE tactile encoder gives a small grasping improvement on Liquid Transfer, likely because this task appears in ViTac-MAE pretraining, but does not yield consistent gains on Chip Pick or Cherry Tomato Pick. Removing the tactile input lowers the average grasp rate by 23.3\%. Removing joint attention lowers the average grasp rate by 18.9\%, showing that tactile and visual features must be jointly modeled for accurate force prediction. Reducing RETAF to 5~Hz causes the largest drop, highlighting the importance of decoupling high-frequency reactive force adaptation from the base policy.}
RETAF also benefits a VLA base policy: with $\pi_{0.5}$~\cite{intelligence2025pi05}, it raises task success from 20\% to 90\% on Cherry Tomato Pick and from 20\% to 50\% on Liquid Transfer (Appendix~\ref{app:vla}).

\begin{figure}[!htbp]
    \centering
    \includegraphics[width=1.0\linewidth]{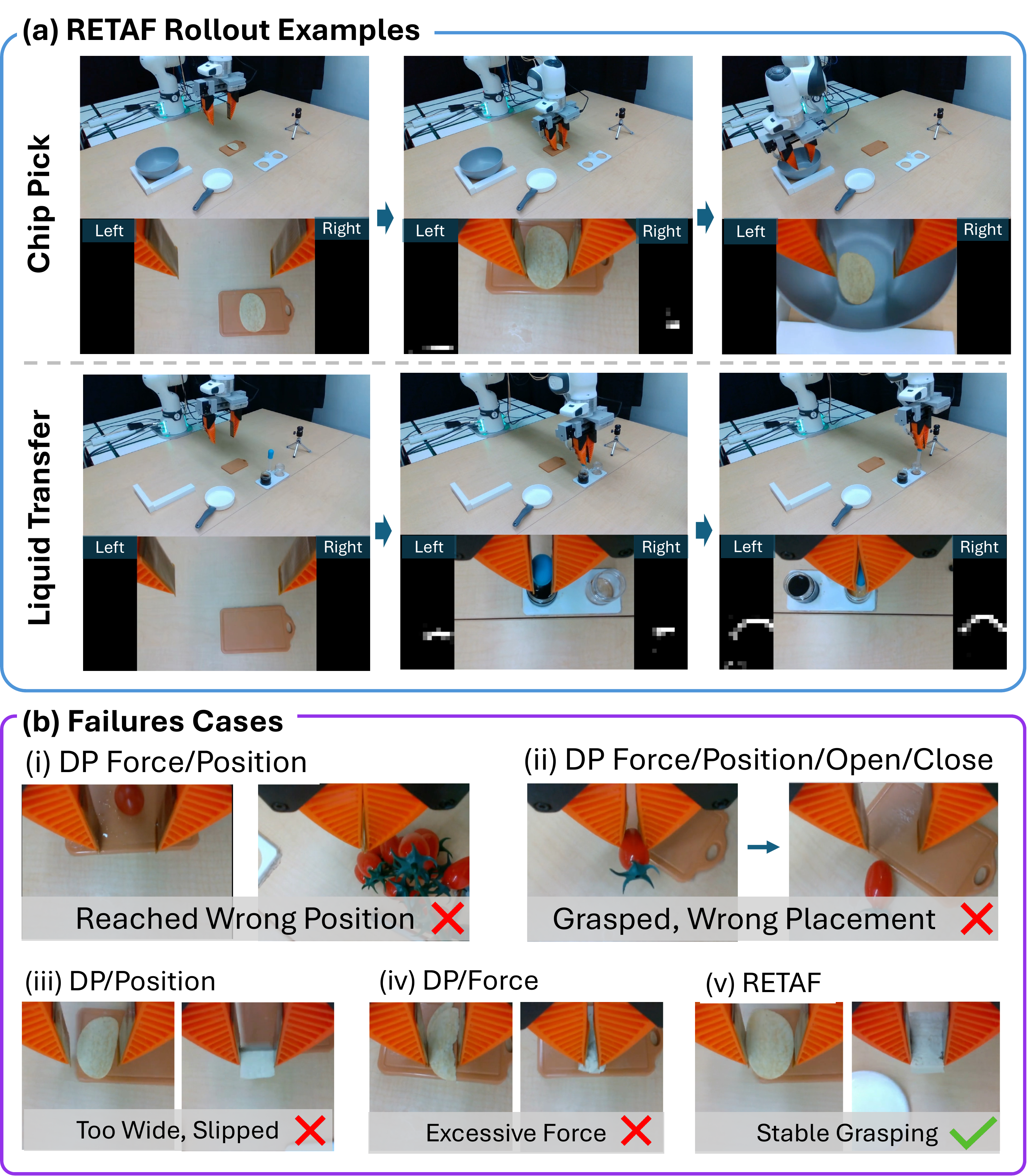}
\caption{
(a) Two successful rollouts, where RETAF correctly predicts appropriate forces and achieves stable grasps. (b) Failure cases: (i) When DP predicts force or position, it shows a higher failure rate during the reach-to-target stage than DP predicting open/close actions. (ii) Regardless of the gripper action type, DP often fails to predict accurate end-effector poses. (iii) We observe that DP--position tends to open too wide, causing slip, while (iv) DP--force tends to predict excessive force, leading to object breakage. In contrast, (v) RETAF predicts more appropriate grasping forces.
}
    \label{fig:qual_examples}
\end{figure}

\subsection{Generalization Studies (RQ5)}
\label{sec:generalization}

\begin{figure}[t]
\centering
\includegraphics[width=\linewidth]{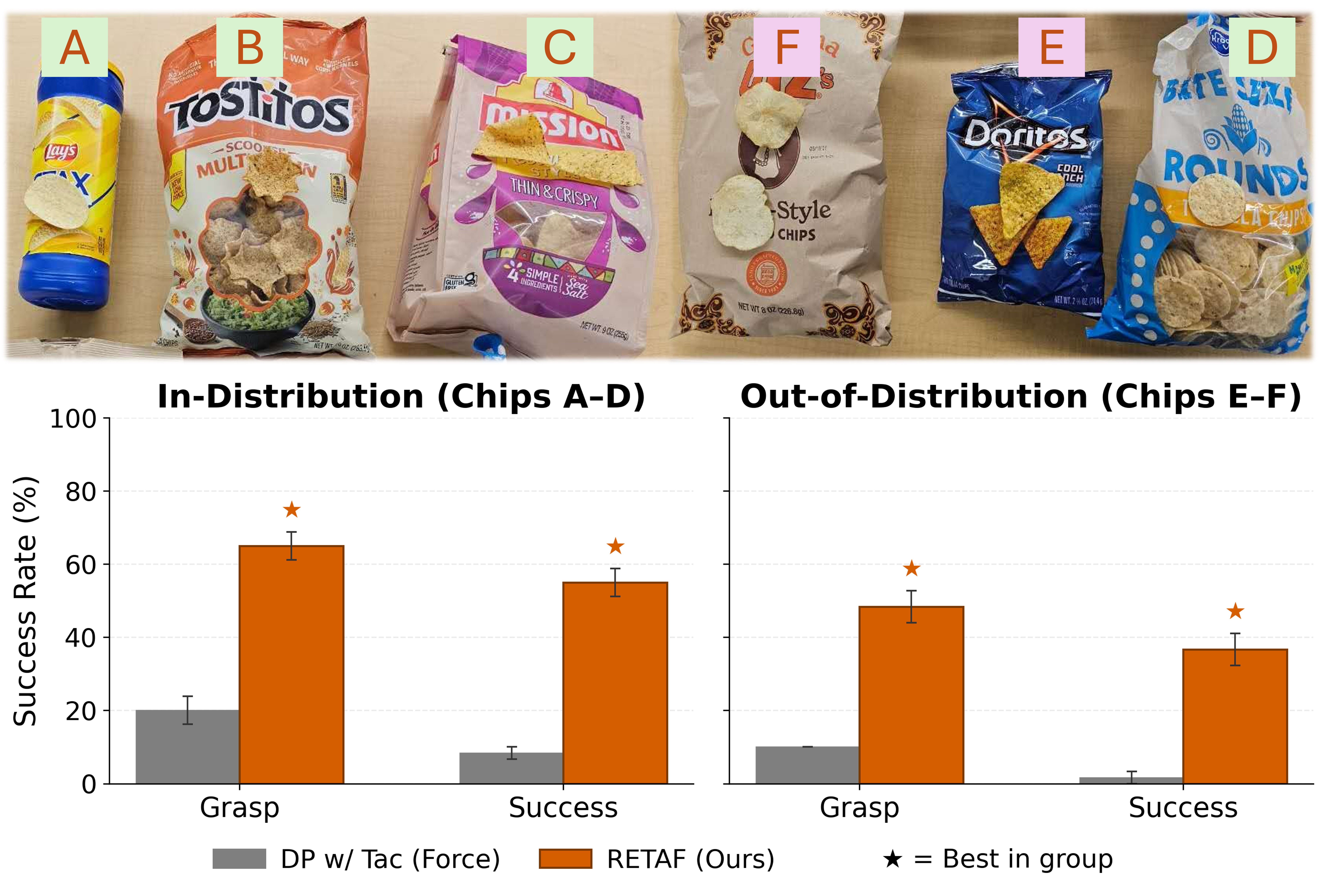}
\caption{{Generalization test on the Chip Pick task. \textbf{Left:}
in-distribution chips (A--D). \textbf{Right:} out-of-distribution chips
(E--F). Bars show mean $\pm$ standard error of Grasp (G) and Success (S)
rates over three seeds, with 10 trials per seed.}}
\label{fig:generalization}
\end{figure}

{To assess object-level generalization, we evaluate Chip Pick on six chip variants: four seen in training (A--D) and two held-out
out-of-distribution (OOD) variants (E--F) with unseen shapes, sizes, and surface properties. We collect 20 demonstrations per seen variant and compare RETAF with DP w/ Tactile, both using grasping force as the action space, with 10 rollouts per chip over three seeds (Fig.~\ref{fig:generalization}).}

{RETAF outperforms DP w/ Tactile on both seen and OOD chips. On OOD chips, DP w/ Tactile almost never succeeds (1.7\%), while RETAF still reaches 36.7\%. This suggests that reactive tactile-force adaptation improves robustness to unseen chip geometries and surface properties within the chip-picking setting. Since this evaluation covers one task category and two held-out chip variants, we interpret it as evidence of object-level generalization.}

{We further evaluate cross-robot transfer of the force-adaptation policy by mounting the same TF-Gripper module on a SO-ARM101~\cite{Knight_Standard_Open_SO-100} arm. The arm motion is teleoperated, while the gripper force is predicted by the same RETAF force policy trained on the Franka chip-picking setup.
Because the arm motion is teleoperated, this experiment isolates the local force-adaptation policy.
Across 20 trials, the transferred system achieves a 55\% grasp success rate, comparable to the Franka setup, suggesting that the local force-adaptation policy transfers across arms when the gripper, tactile sensor, and wrist camera are kept as a common end-effector module.}

{In a cross-gripper test on the Franka Hand under position control (supported by both grippers), RETAF improves over DP on both TF-Gripper (7/20 to 13/20) and the Franka Hand (4/20 to 11/20) on Cherry Tomato Pick, showing that the gain of RETAF does not depend on our hardware (Appendix~\ref{app:cross_gripper}).}

{\subsection{Rigid-Object Contact-Rich Tasks}
To examine whether TF-Gripper is confined to delicate objects, we also include two rigid-object contact-rich tasks, \emph{Peg Insertion} and \emph{Block Insertion} (inserting a cube into a matching recess), with 50 force-control demonstrations each (Fig.~\ref{fig:rigid_tasks}).
RETAF and DP w/ Force reach similar success rates over 30 rollouts, since a rigid object tolerates a wide range of grasping forces and success depends mainly on the pose precision of the insertion.
While such arm-level contact-rich manipulation~\cite{xue2025reactive,hou2024adaptive} is not the target of RETAF, these tasks, together with everyday tasks without force regulation (Appendix~\ref{app:rigid_everyday}), show that the TF-Gripper itself can apply to diverse tasks.
\begin{figure}[b]
\centering
\includegraphics[width=\linewidth]{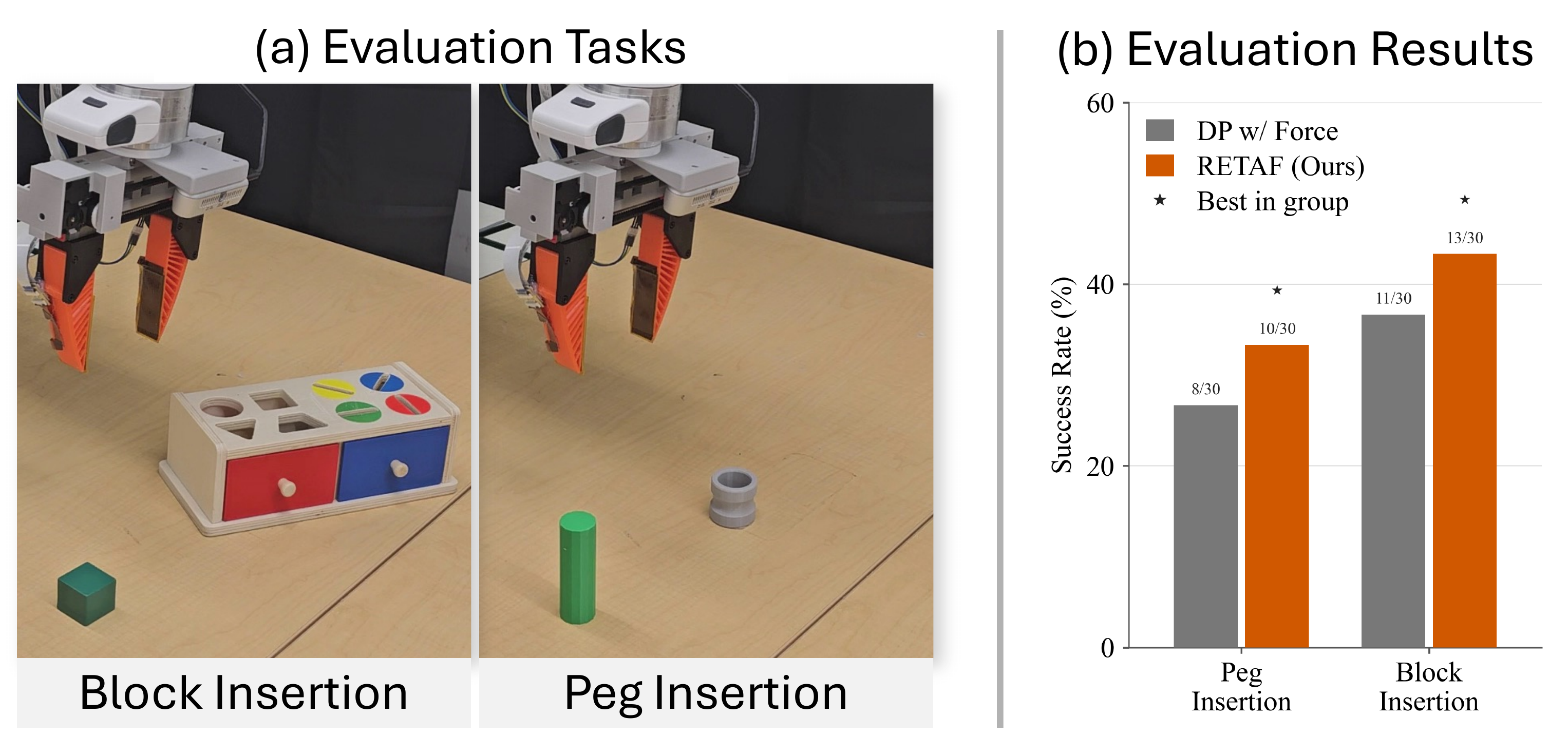}
\caption{{Rigid-object contact-rich tasks. (a) Block Insertion and Peg Insertion. (b) Success rates over 30 rollouts per task.}}
\label{fig:rigid_tasks}
\end{figure}
}

{\subsection{Data-Scaling Analysis}
We train RETAF with 10, 25, 50, and 100 demonstrations on Cherry Tomato Pick and Peg Insertion, with 30 rollouts per setting (Table~\ref{tab:data_scaling}). These are independent runs from the main experiments in Fig.~\ref{fig:five_mani_tasks_results}.
On Cherry Tomato Pick, the model reaches reasonable performance with 50 demonstrations (73.3\% success), and 100 demonstrations improve it only slowly (+6.7\%).
On Peg Insertion, the grasp rate is already 83.3\% at 50 demonstrations, so the remaining failures mainly come from the insertion, which needs precise pose prediction from the base policy; since more demonstrations mainly improve pose prediction, task success keeps rising (33.3\% to 46.7\%) while the grasp rate changes little.
\begin{table}[t]
\centering
\caption{{Data scaling of RETAF: Grasp\,/\,Success rates (\%) vs.\ number of demonstrations, 30 rollouts per setting.}}
\label{tab:data_scaling}
{\scriptsize
\renewcommand{\arraystretch}{0.95}
\setlength{\tabcolsep}{4pt}
\resizebox{0.90\linewidth}{!}{%
\begin{tabular}{@{}lcccc@{}}
\toprule
Task & 10 demos & 25 demos & 50 demos & 100 demos \\
\midrule
Cherry Tomato Pick & 30.0\,/\,6.7 & 73.3\,/\,36.7 & \cellcolor{ourshl!40}83.3\,/\,73.3 & 90.0\,/\,80.0 \\
Peg Insertion & 36.7\,/\,6.7 & 66.7\,/\,16.7 & \cellcolor{ourshl!40}83.3\,/\,33.3 & 96.7\,/\,46.7 \\
\bottomrule
\end{tabular}%
}}
\end{table}
}

\subsection{Qualitative Examples}

Fig.~\ref{fig:qual_examples}(a) provides two qualitative rollout examples of RETAF, where RETAF correctly predicts appropriate forces and achieves stable grasps.
Fig.~\ref{fig:qual_examples}(b) illustrates common failure cases. During the reach-to-target stage, DP that predicts positions or forces exhibits a higher failure rate than DP predicting open/close actions; however, predicting open/close does not eliminate failures. In some cases, the grasp succeeds, but the task still fails due to incorrect pose predictions by DP.
During grasping, we observe that DP predicting position tends to result in insufficient contact and slipping, consistent with our earlier analysis that position control is less robust to small variations in object width and can cause the policy to assume contact falsely.
In contrast, DP with direct force prediction tends to output excessive forces. Force control can tolerate slightly smaller forces while still achieving task success, consistent with our earlier analysis of its robustness. However, without timely reactions to tactile feedback, DP often overestimates the required force. Overall, RETAF predicts more accurate and well-regulated grasping forces.

\section{Conclusion and Future Work}

In this work, we investigate how explicit force control with tactile feedback benefits learning-based manipulation.
We develop TF-Gripper with a force-aware teleoperation device for data collection, and RETAF, a policy framework that decouples force regulation from pose prediction.
Across six real-world tasks, force control consistently outperforms position control in grasp stability, and RETAF further improves task performance across base policies.
Our evaluation is empirical and limited to the selected tasks.
Future work includes analyzing force- versus position-based control theoretically, extending the evaluation to new task categories for task-level generalization, and scaling force-regulated manipulation with large-scale force and tactile data.
\section*{Acknowledgments}
This work was partly supported by Delta Electronics Inc.
We acknowledge Research Computing at the University of Virginia for providing the computational resources that made the results in this work possible.

\bibliography{reference}
\bibliographystyle{IEEEtran}

\clearpage
\appendix

\subsection{Force-Control Accuracy of TF-Gripper}
\label{app:force_accuracy}

\textbf{Low-force accuracy and hysteresis.}
The PWM--force mapping of TF-Gripper is calibrated with a force gauge at the fingertip (Fig.~\ref{fig:pwm_force}), so static friction in the transmission is already included in the calibration.
The remaining nonlinearity is the direction-dependent hysteresis of the transmission.
To quantify it, we measure the fingertip force with the force gauge at commanded forces of 1, 5, and 10~N, during loading (1$\rightarrow$5$\rightarrow$10~N) and during unloading from 10~N (10$\rightarrow$5$\rightarrow$1~N), with 10 repetitions each.
Table~\ref{tab:app_force_accuracy} shows the results.
During loading, the measured force is within 5\% of the command at all three setpoints.
With only the loading-direction calibration, unloading shows the hysteresis: the measured force at a commanded 5~N drops to 4.06~N, an error of about 1~N (19\%).
We therefore calibrate the loading and unloading directions separately.
With this bidirectional calibration, the mean error is within 7\% at all setpoints, which is on par with the $\pm$10\% force repeatability specified for the Robotiq 2F-85~\cite{robotiq2019gripper}.

\begin{table}[h]
\centering
\caption{Measured fingertip force of TF-Gripper (mean $\pm$ std over 10 repetitions). The middle group uses only the loading-direction calibration to show the hysteresis. The bottom group uses the bidirectional calibration that TF-Gripper uses in all experiments.}
\label{tab:app_force_accuracy}
\small
\setlength{\tabcolsep}{5pt}
\begin{adjustbox}{max width=\linewidth}
\begin{tabular}{@{}lcc@{}}
\toprule
Condition & Commanded (N) & Measured (N) \\
\midrule
Loading (1$\rightarrow$5$\rightarrow$10~N) & 1 & $0.95 \pm 0.06$ \\
 & 5 & $5.06 \pm 0.47$ \\
 & 10 & $10.22 \pm 0.88$ \\
\midrule
Unloading (10$\rightarrow$5$\rightarrow$1~N), & 5 & $4.06 \pm 0.92$ \\
loading-only calibration & 1 & $0.81 \pm 0.19$ \\
\midrule
\rowcolor{ourshl!40} Unloading (10$\rightarrow$5$\rightarrow$1~N), & 5 & $4.98 \pm 0.65$ \\
\rowcolor{ourshl!40} bidirectional calibration & 1 & $1.07 \pm 0.08$ \\
\bottomrule
\end{tabular}
\end{adjustbox}
\end{table}

\textbf{Comparison with the Franka Hand at high forces.}
The Franka Hand cannot apply forces below 30~N~\cite{franka2020hand}, so it cannot be compared with TF-Gripper in the force range of our tasks (0.5--7~N, Table~\ref{tab:task_setup}).
Its force range overlaps with that of TF-Gripper (0.45--45~N) above 30~N, so we compare the two grippers in this overlapping range.
We command 32~N and 35~N on each gripper and measure the fingertip force with the same force gauge, with 10 repetitions per setting (Table~\ref{tab:app_highforce}).
The Franka Hand has a larger mean bias (about 1~N below the command) and a smaller standard deviation.
TF-Gripper has a smaller mean bias (within 0.4~N) and a larger standard deviation.
In this range, the force accuracy of the two grippers is comparable.

\begin{table}[h]
\centering
\caption{Measured fingertip force (mean $\pm$ std over 10 repetitions) of TF-Gripper and the Franka Hand at commanded forces in their overlapping range.}
\label{tab:app_highforce}
\small
\setlength{\tabcolsep}{6pt}
\begin{tabular}{@{}ccc@{}}
\toprule
Commanded (N) & TF-Gripper (N) & Franka Hand (N) \\
\midrule
32 & $32.36 \pm 1.07$ & $31.15 \pm 0.55$ \\
35 & $35.12 \pm 1.14$ & $33.94 \pm 0.68$ \\
\bottomrule
\end{tabular}
\end{table}

\textbf{Belt load.}
The transmission is a toothed HTD-3M timing belt that engages toothed pulleys.
The grasping force is transmitted through tooth meshing, not through belt friction.
The admissible working tension is approximately 100~N for the 10~mm-wide belt used in TF-Gripper~\cite{maedler_htd3m}.
The forces in our tasks are at most 10\% of this value, and the 45~N maximum output of the gripper stays below half of it.
The belt therefore works in a small part of its rated range.
We expect the residual friction and hysteresis to come mainly from the gear train of the actuators and the linear guides, which is what the measurements in Table~\ref{tab:app_force_accuracy} quantify.

\subsection{Force and Tactile Dynamics Around Contact}
\label{app:contact_dynamics}
\begin{figure}[h]
    \centering
    \includegraphics[width=1.0\linewidth]{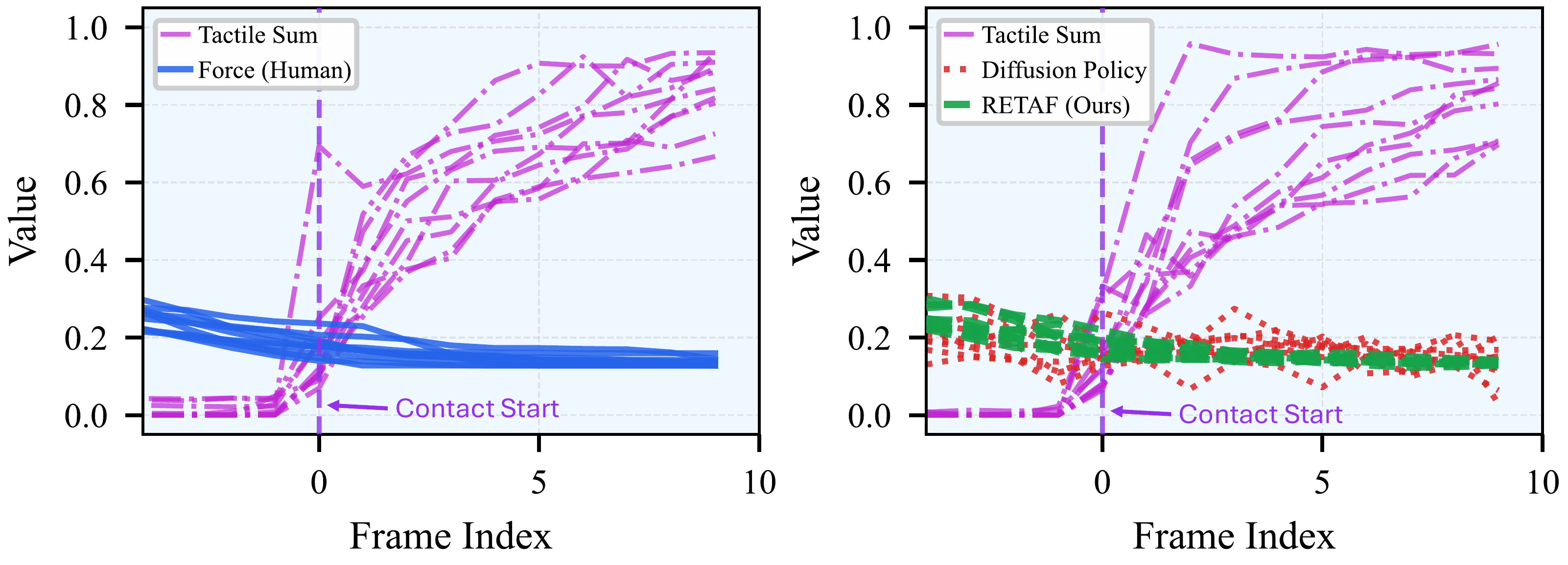}
\caption{Force and tactile signals before and after contact in human demonstrations (left) and in rollouts (right), for 10 randomly selected trajectories of Cherry Tomato Pick. The tactile sum is min-max scaled to $[0, 1]$. The force is scaled to $[-1, 1]$, where values above 0 give the magnitude of the grasping force.}
    \label{fig:contact_analysis}
\end{figure}
We analyze how the force command changes with the tactile signal around the first contact on Cherry Tomato Pick.
At this moment, the gripper goes from free motion to physical interaction, and the force must be adjusted quickly.
Fig.~\ref{fig:contact_analysis} shows the force commands aligned with the tactile feedback in human demonstrations and in policy rollouts.
In human demonstrations, the force is adjusted right after the tactile signal rises.
In rollouts, DP follows the overall trend, but its predictions are noisier.
The force predicted by RETAF follows the pattern of the human demonstrations more closely.
\begin{figure}[t]
    \centering
    \includegraphics[width=1.0\linewidth]{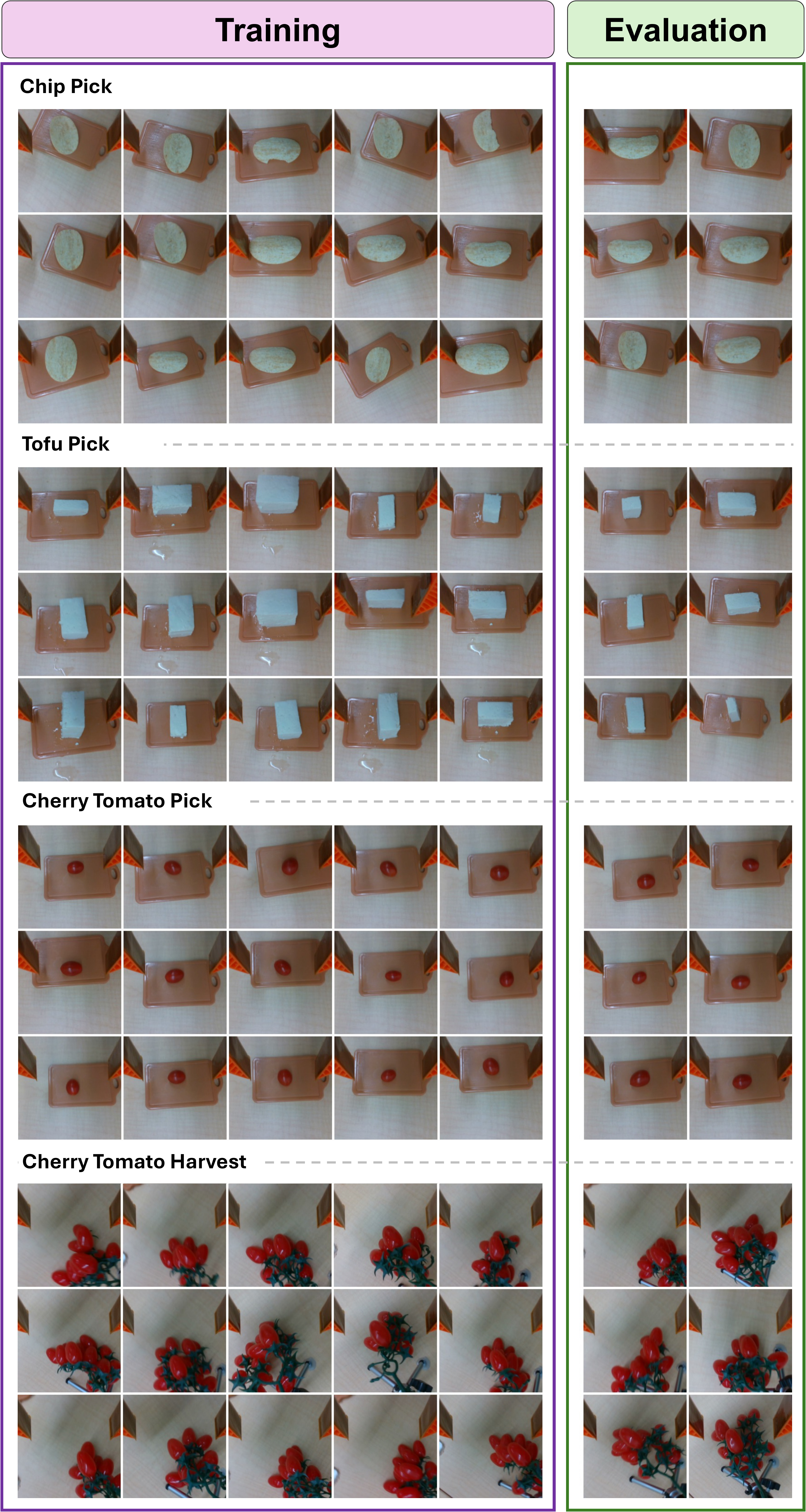}
\caption{Randomly selected wrist-view images captured before grasping, for four tasks. For each task, we show 15 training samples and 6 rollout frames. Liquid Transfer is not shown because the cup and the tube are mounted on a fixed rig, and its wrist views are largely identical.}
    \label{fig:data_vis}
\end{figure}

\subsection{Data Diversity in Training and Rollouts}
\label{app:data_diversity}

Fig.~\ref{fig:data_vis} shows randomly sampled wrist-view images captured before grasping, from the training data and from the evaluation rollouts.
The rollouts cover a similar distribution of object poses and appearances as the training data.
The comparison between RETAF and the baselines in the main experiments is therefore made under in-distribution conditions.

\subsection{RETAF with a VLA Base Policy}
\label{app:vla}

Table~\ref{tab:app_vla} gives the per-stage results of pairing RETAF with $\pi_{0.5}$~\cite{intelligence2025pi05} as the base policy.
Each setting is evaluated with 10 rollouts.
Without RETAF, $\pi_{0.5}$ reaches the object in most rollouts but often fails to grasp it stably.
With RETAF, the stable grasp rate and the task success rate increase on both tasks.

\begin{table}[h]
\centering
\caption{RETAF with $\pi_{0.5}$ as the base policy. Tasks are evaluated in three stages: Reach (R), Stable Grasp (G), and Task Success (S), with 10 rollouts per setting.}
\label{tab:app_vla}
\small
\setlength{\tabcolsep}{5pt}
\begin{tabular}{@{}lcccccc@{}}
\toprule
& \multicolumn{3}{c}{Cherry Tomato Pick} & \multicolumn{3}{c}{Liquid Transfer} \\
\cmidrule(lr){2-4} \cmidrule(l){5-7}
Method & R & G & S & R & G & S \\
\midrule
$\pi_{0.5}$ & 80\% & 30\% & 20\% & 60\% & 20\% & 20\% \\
\rowcolor{ourshl!40} $\pi_{0.5}$ + RETAF & 100\% & 90\% & 90\% & 70\% & 60\% & 50\% \\
\bottomrule
\end{tabular}
\end{table}

\subsection{Can RETAF Replace Gripper Action Prediction in the Base Policy?}
\label{app:gripper_action}

We study whether the Force Adaptation Policy can also decide when to grasp.
We compare the standard RETAF setting with a variant in which the base policy predicts only the arm pose, and the Force Adaptation Policy predicts both the open/close action and the force.
As shown in Table~\ref{tab:app_gripper_action}, removing the open/close prediction from the base policy leads to a large performance drop.
Without the global view, the Force Adaptation Policy sees only the wrist image and the tactile feedback, and it struggles to infer the task progress and to decide when to grasp.
These results suggest that deciding when to grasp and regulating the force during contact are two different roles.
The first needs the global task context of the base policy, and the second needs fast local feedback.

\begin{table}[h]
\centering
\caption{Ablation on removing the open/close prediction from the base policy, with 10 rollouts per setting. These single-run results are not directly comparable to the three-seed results in Fig.~\ref{fig:five_mani_tasks_results}.}
\label{tab:app_gripper_action}
\resizebox{\columnwidth}{!}{
\begin{tabular}{lcccccc}
\toprule
& \multicolumn{3}{c}{RETAF}
& \multicolumn{3}{c}{RETAF w/o open/close in base policy} \\
\cmidrule(lr){2-4} \cmidrule(lr){5-7}
Task
& Reach
& \begin{tabular}[c]{@{}c@{}}Stable\\Grasp\end{tabular}
& \begin{tabular}[c]{@{}c@{}}Task\\Success\end{tabular}
& Reach
& \begin{tabular}[c]{@{}c@{}}Stable\\Grasp\end{tabular}
& \begin{tabular}[c]{@{}c@{}}Task\\Success\end{tabular} \\
\midrule
Cherry Tomato Pick
& 100\% & 90\% & 80\%
& 100\% & 20\% & 0\% \\
Liquid Transfer
& 70\% & 60\% & 60\%
& 80\% & 0\% & 0\% \\
\bottomrule
\end{tabular}
}
\end{table}

\subsection{Rigid-Object and Everyday Tasks}
\label{app:rigid_everyday}

Table~\ref{tab:app_rigid} gives the success counts behind Fig.~\ref{fig:rigid_tasks}(b).
For each task, we collect 50 force-control demonstrations and evaluate each policy with 30 rollouts.
Both policies complete the two tasks, and their success rates are similar.
A rigid object can be grasped with a wide range of forces, so success depends mainly on the pose precision of the insertion.
The absolute success rates are modest because both tasks have tight insertion tolerances.

\begin{table}[h]
\centering
\caption{Task success on the two rigid-object tasks with TF-Gripper, over 30 rollouts per task.}
\label{tab:app_rigid}
\small
\setlength{\tabcolsep}{6pt}
\begin{tabular}{@{}lcc@{}}
\toprule
Method & Peg Insertion & Block Insertion \\
\midrule
DP w/ Force & 8/30 (26.7\%) & 11/30 (36.7\%) \\
\rowcolor{ourshl!40} RETAF & 10/30 (33.3\%) & 13/30 (43.3\%) \\
\bottomrule
\end{tabular}
\end{table}

\begin{figure}[h]
\centering
\includegraphics[width=0.8\linewidth]{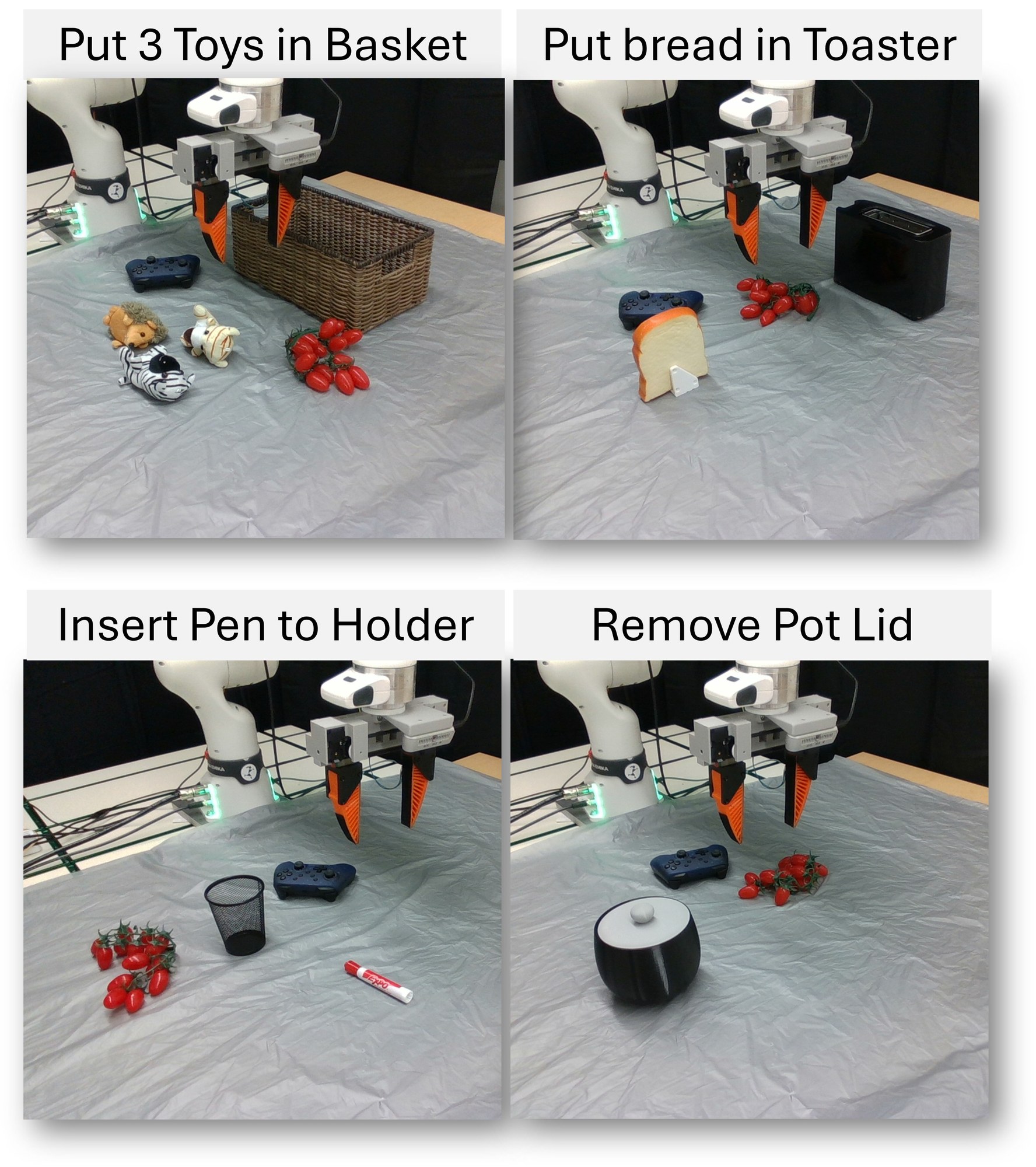}
\caption{Everyday tasks performed with TF-Gripper.}
\label{fig:app_everyday}
\end{figure}

TF-Gripper also works for everyday tasks that do not require force regulation.
Fig.~\ref{fig:app_everyday} shows four examples: putting toys in a basket, putting bread in a toaster, inserting a pen into a holder, and removing a pot lid.
These tasks are run with a fixed gripper position or a fixed grasping force, without tactile feedback.

\subsection{Cross-Gripper Comparison with the Franka Hand}
\label{app:cross_gripper}

\begin{figure}[h]
\centering
\includegraphics[width=\linewidth]{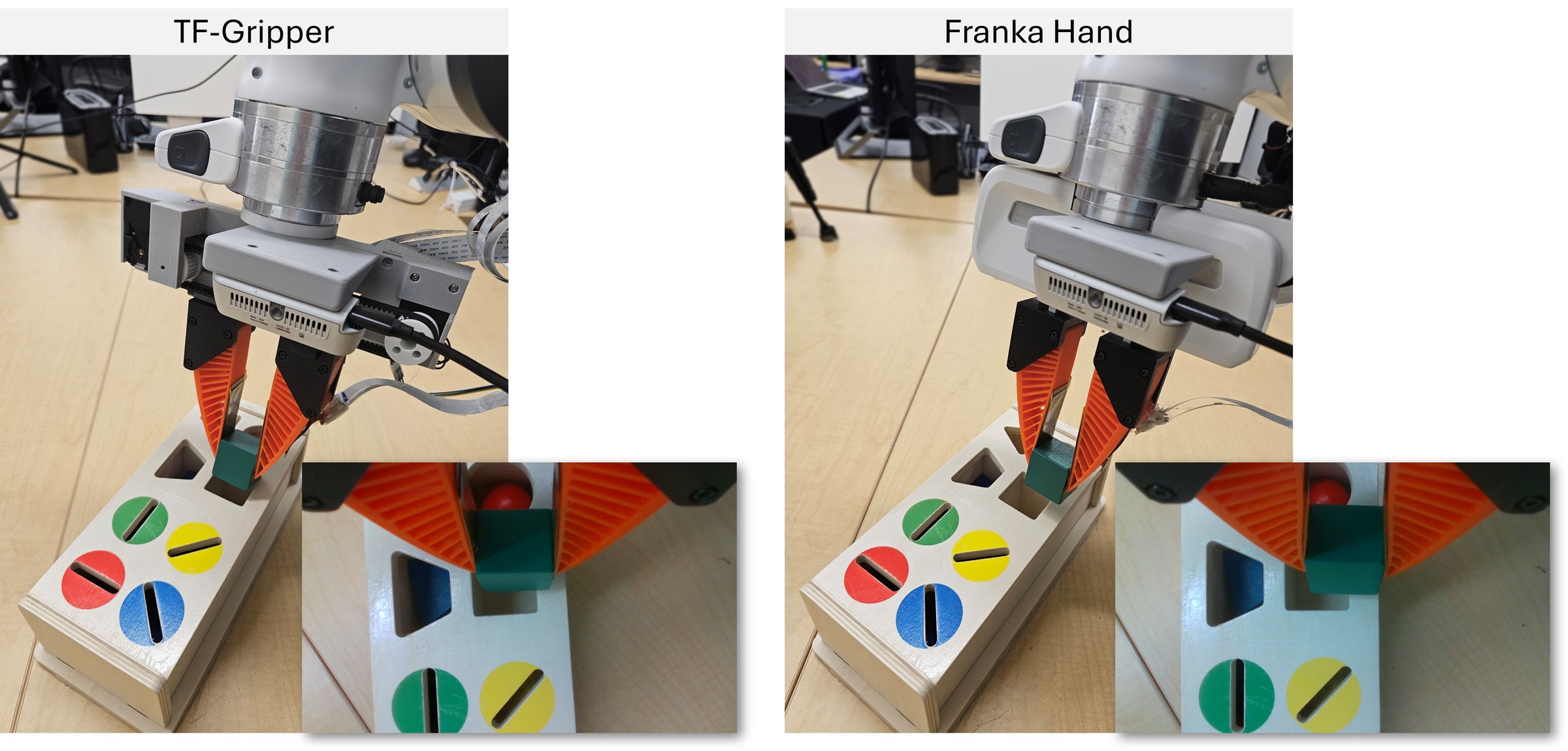}
\caption{Cross-gripper setup, shown with the Block Insertion task. TF-Gripper (left) and the Franka Hand (right) are mounted on the same Franka arm with the same tool center point, wrist camera, and tactile fingertips. The insets show the wrist-camera views.}
\label{fig:app_cross_gripper}
\end{figure}

We test whether the gain of RETAF depends on our hardware by running the same policies on the Franka Hand.
A different gripper requires new demonstrations, so a perfectly controlled comparison is not possible.
We reduce the other differences as much as we can.
Both grippers are mounted on the same Franka arm with the same tool center point, the same wrist camera, and the same fingertips with tactile sensors, so the wrist views are nearly identical (Fig.~\ref{fig:app_cross_gripper}).
Since the Franka Hand cannot regulate small forces, both grippers use position control.
In this setting, the Force Adaptation Policy of RETAF predicts the gripper position instead of the force.

\begin{table}[h]
\centering
\caption{Cross-gripper comparison on Cherry Tomato Pick under position control, which both grippers support. Values are task success over 20 trials per setting.}
\label{tab:app_cross_gripper}
\small
\setlength{\tabcolsep}{6pt}
\begin{tabular}{@{}lcc@{}}
\toprule
Method & TF-Gripper & Franka Hand \\
\midrule
DP & 7/20 (35.0\%) & 4/20 (20.0\%) \\
\rowcolor{ourshl!40} RETAF & 13/20 (65.0\%) & 11/20 (55.0\%) \\
\bottomrule
\end{tabular}
\end{table}

Table~\ref{tab:app_cross_gripper} shows the results on Cherry Tomato Pick with 20 trials per setting.
RETAF improves over DP on both grippers, from 7/20 to 13/20 on TF-Gripper and from 4/20 to 11/20 on the Franka Hand.
This suggests that the gain of RETAF is not tied to our hardware.
TF-Gripper is slightly better than the Franka Hand for both policies.
The most noticeable difference between the two grippers in these experiments is the lower actuation latency of TF-Gripper (about 50~ms vs.\ 70--100~ms for the Franka Hand), which may explain this gap.
We also run DP on Block Insertion with position control on both grippers.
DP succeeds in 11/20 trials on TF-Gripper and in 9/20 trials on the Franka Hand.

\end{document}